\documentclass[10pt,twocolumn,letterpaper]{article}

 \usepackage[pagenumbers]{iccv} 

\usepackage{graphicx}
\usepackage{amsmath,amssymb} 
\usepackage{caption} 
\usepackage{wrapfig}
\usepackage{csquotes}
\usepackage{bm}
\usepackage{paralist,algorithm}
\usepackage[noend]{algpseudocode}
\usepackage{multirow}

\newcommand{\x}{{\bf x}}

\newcommand{\p}{{\bf p}}

\newcommand{\w}{{\bf w}}

\newcommand{\z}{{\bf z}}

\newcommand{\D}{\mathcal{D}}

\newcommand{\R}{\mathbb{R}}

\newcommand{\name}{{\sc Engine }}
\newcommand{\mame}{{\sc Engine}}

\usepackage{accents}
\newlength{\dhatheight}

\usepackage{xcolor,colortbl}
\definecolor{Gray}{gray}{0.85}
\definecolor{Redo}{rgb}{0.95,0.69,0.51}
\definecolor{LightCyan}{rgb}{0.88,1,1}
\usepackage{afterpage}
\usepackage{pifont}

\usepackage{csquotes}

\usepackage{lipsum}
\definecolor{cvprblue}{rgb}{0.21,0.49,0.74}
\usepackage[misc]{ifsym}

\definecolor{iccvblue}{rgb}{0.21,0.49,0.74}
\usepackage[pagebackref,breaklinks,colorlinks,allcolors=iccvblue]{hyperref}

\newcommand{\blue}[1]{{\color{blue}#1}}
\title{External Knowledge Injection for CLIP-Based Class-Incremental Learning}

\author{Da-Wei Zhou$^{1,2}$, Kai-Wen Li$^{1,2}$, Jingyi Ning$^{2}$, Han-Jia Ye$^{1,2}$\textsuperscript{(\Letter)}, Lijun Zhang$^{1,2}$, De-Chuan Zhan$^{1,2}$\\
	$^{1} $ School of Artificial Intelligence, Nanjing University\\
	$^{2} $ National Key Laboratory for Novel Software Technology, Nanjing University\\
	{\tt\small {\{zhoudw,likw,yehj,zhanglj,zhandc\}@lamda.nju.edu.cn}, ningjy@nju.edu.cn}
}

\begin{document}
\maketitle
\footnotetext[2]{Correspondence to: Han-Jia Ye (yehj@lamda.nju.edu.cn)}
\begin{abstract}

Class-Incremental Learning (CIL) enables learning systems to continuously adapt to evolving data streams. With the advancement of pre-training, leveraging pre-trained vision-language models (\eg, CLIP) offers a promising starting point for CIL. 
However, CLIP makes decisions by matching visual embeddings to class names, overlooking the rich contextual information conveyed through language. 
For instance, the concept of ``cat'' can be decomposed into features like tail, fur, and face for recognition.
Besides, since the model is continually updated, these detailed features are overwritten in CIL, requiring external knowledge for compensation.
In this paper, we introduce ExterNal knowledGe INjEction (\mame) for CLIP-based CIL. To enhance knowledge transfer from outside the dataset, we propose a dual-branch injection tuning framework that encodes informative knowledge from both visual and textual modalities. The visual branch is enhanced with data augmentation to enrich the visual features, while the textual branch leverages GPT-4 to rewrite discriminative descriptors. In addition to this on-the-fly knowledge injection, we also implement post-tuning knowledge by re-ranking the prediction results during inference. With the injected knowledge, the model can better capture informative features as data evolves. Extensive experiments demonstrate its state-of-the-art performance. Code is available at: \url{https://github.com/LAMDA-CL/ICCV25-ENGINE}.

\end{abstract}    
\section{Introduction} \label{sec:intro}

\begin{figure}[t]
	\vspace{-6mm}
	\begin{center}
		{\includegraphics[width=.95\columnwidth]{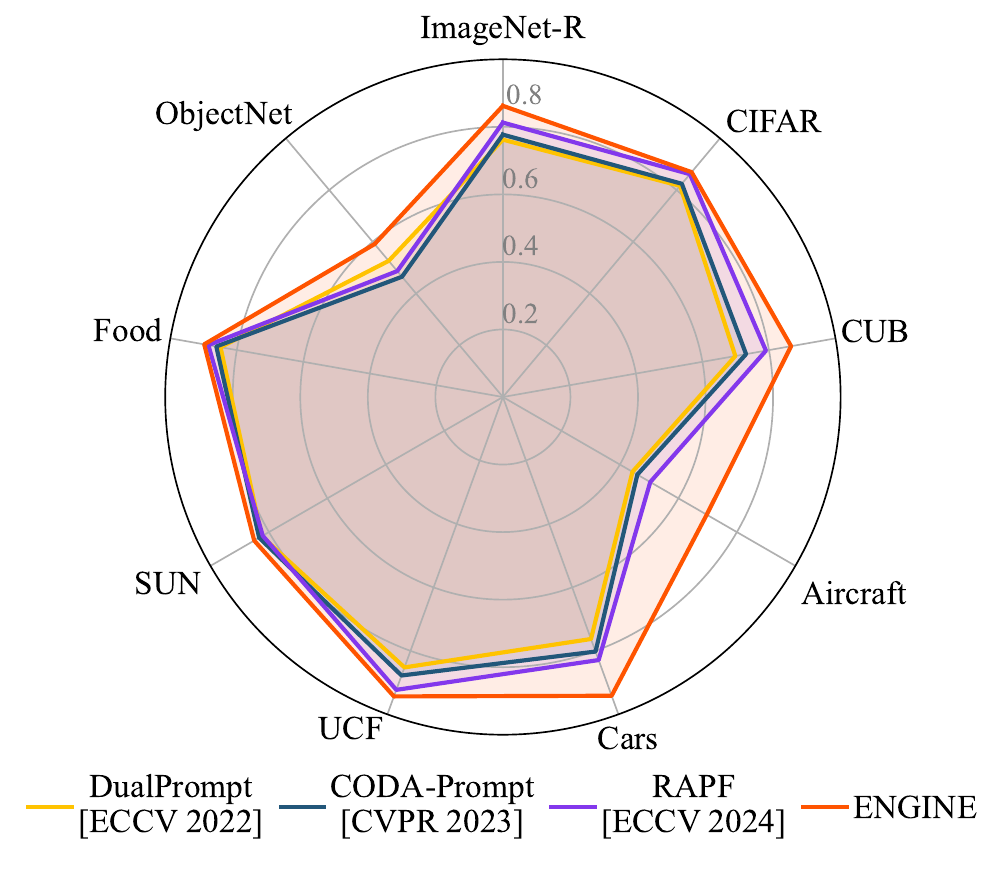}}
	\end{center}
	\vspace{-7mm}
	\caption{\small Last accuracy comparison on nine benchmark datasets. All methods are deployed with the same CLIP for fair comparison. \name outperforms other competitors with a substantial margin.
	}
	\vspace{-7mm}
	\label{figure:radar}
\end{figure}

Recent advancements in deep learning have significantly impacted various aspects of life~\cite{deng2009imagenet,he2015residual,yang2023exploring}. However, real-world data often presents substantial challenges to these models, particularly streaming data that requires continual learning~\cite{aggarwal2018survey,li2024configure}. Class-Incremental Learning (CIL)~\cite{rebuffi2017icarl} has been proposed to address this challenge by enabling models to absorb new knowledge incrementally. A major issue in CIL is catastrophic forgetting~\cite{french1999catastrophic,french1999modeling}, where learning new classes leads to the loss of previously acquired knowledge. As a result, much research has focused on mitigating forgetting in deep learning~\cite{masana2022class,de2021survey}.
In contrast to traditional approaches that train models from scratch, recent progress in pre-training has shifted attention towards using pre-trained models (PTMs)~\cite{wang2022learning}, such as Vision Transformers\cite{dosovitskiy2020image} and CLIP~\cite{radford2021learning}. With extensive training data serving as pre-learned knowledge, PTMs offer a more generalizable starting point for CIL, showing promise for real-world applications. \looseness=-1

CLIP~\cite{radford2021learning}, as a pioneering pre-trained vision-language model~\cite{jia2021scaling,yang2023learning,yu2022coca,yuan2021florence,li2023blip}, exhibits strong zero-shot performance by matching the embeddings of query images to class names. 
However, since visual features often contain fine-grained information~\cite{menon2022visual}, these detailed descriptors are neglected when using class names as matching targets.
For example, the text concept of ``a photo of a cat'' can be further decomposed into features like ``whiskers,'' ``tail,'' and ``fur'' in order to match the visual embeddings, while these informative features are not well utilized. Furthermore, as the model is updated in CIL, the sub-features can be overwritten during the learning process, resulting in forgetting and a mismatch between cross-modal features.  \looseness=-1

Recent studies~\cite{wang2022learning,wang2023attriclip,zhou2023learning,yi2024leveraging} have highlighted the limitations of using fixed template text in vision-language models, which often leads to a loss of rich contextual information. To address this, researchers have proposed learnable text inputs, such as task-specific prompts~\cite{zhou2022learning}, which allow the model to encode additional, task-relevant information. However, these learnable inputs can become biased towards the visual features in the training set, ultimately compromising the model's generalizability~\cite{zhou2022conditional}.

In CLIP, a more detailed description in the textual space can provide a more informative matching target, enhancing prediction accuracy. For example, instead of using the generic template ``a photo of a cat,'' directly describing its features, such as ``whiskers'' and ``tail,'' can significantly improve recognition.
In this way, these detailed features serve as anchors in the embedding space, and aligning to them helps reduce the risk of catastrophic forgetting.

Learning with informative descriptors offers a promising approach to enhancing CLIP's continual learning capabilities. However, two major challenges remain: {\bf 1)} acquiring informative features. Given the diversity of classes in downstream tasks, it is impractical for human experts to annotate all relevant features for recognition. {\bf 2)} mitigating forgetting when utilizing external knowledge. As CIL involves continuous data streams, the model risks forgetting previously acquired knowledge after each update. Therefore, it is crucial to efficiently extract and retain external knowledge while preventing forgetting during sequential updates.

To address these challenges, we propose ExterNal knowledGe InjEction (\mame) for CLIP-based CIL.
To acquire informative features, we enhance knowledge transfer by incorporating external knowledge from both visual and textual modalities through a dual-branch tuning framework. The visual branch is enhanced via data augmentation to enrich visual features, while the textual branch uses GPT-4 to rewrite discriminative prompts. In this way, we can encode diverse and informative class descriptions in the continual updating process.
 In addition to on-the-fly knowledge injection, we also apply post-tuning by re-ranking predictions during inference. Post-tuning is done by considering the local pair-wise features, which can further calibrate the wrong predictions to boost incremental learning.

\section{Related Work}

\noindent\textbf{Class-Incremental Learning (CIL)}: is a long-standing problem, aiming to absorb new knowledge without forgetting~\cite{masana2022class,de2021survey}. Typical CIL algorithms train a model from scratch, which can be divided into several groups.
Knowledge distillation-based methods aim to build the mapping target~\cite{hinton2015distilling} between old and new models to resist forgetting~\cite{rebuffi2017icarl,li2016learning,douillard2020podnet}, which involve logit-wise alignment~\cite{rebuffi2017icarl,li2016learning}, feature-wise alignment~\cite{hou2019learning,lu2022augmented,park2021class}, and group-wise alignment~\cite{gao2022rdfcil,tao2020topology,dong2021few}. Replay-based methods aim to recover previous knowledge by saving and replaying the subset of seen classes~\cite{luo2023class,aljundi2019gradient,chaudhry2018riemannian,liu2020mnemonics,chaudhry2018efficient,zhengmulti}.
Besides, parameter regularization-based methods estimate the importance of parameters and restrict important ones not to change~\cite{kirkpatrick2017overcoming,aljundi2018memory,zenke2017continual,aljundi2019task}.
Model rectification-based methods observe and rectify the inductive bias in the model, \eg, the biased logits~\cite{shi2022mimicking,zhao2020maintaining,yu2020semantic} and classifier weights~\cite{zhao2020maintaining,wu2019large}. Model expansion-based methods~\cite{yan2021dynamically,wang2022foster,zhou2022model} adjust the network structure to fit the characteristics of the evolving data, \eg, expanding neurons~\cite{yoon2018lifelong,xu2018reinforced}, the whole backbone~\cite{yan2021dynamically,wang2022foster,zhou2022model,wang2023beef,zheng2025task}, and lightweight modules~\cite{douillard2022dytox}.

\noindent\textbf{Pre-Trained Model-Based CIL}: improves the capability of CIL by starting with generalizable pre-trained models~\cite{zhou2025duct,qi2025adaptive,sun2024mos,li2025addressing}. To maintain the pre-trained knowledge, most works seek to freeze the backbone and append lightweight modules, \eg, prompts~\cite{wang2022dualprompt,wang2022learning,smith2023coda,wang2022s,zhou2022learning} and adapters~\cite{rebuffi2017learning,yu2024boosting,chen2022adaptformer}. For example, L2P~\cite{wang2022learning} and DualPrompt~\cite{wang2022dualprompt} design the visual prompt~\cite{jia2022visual} pool to select the instance-specific prompts when learning with the pre-trained Vision Transformer. Several works build complex prompt combination/generation targets with attention mechanism~\cite{smith2023coda} or generative network~\cite{jung2023generating}. Other works directly utilize the pre-trained features to construct strong classifiers by matching class prototypes to the embeddings~\cite{zhou2023revisiting,mcdonnell2023ranpac,snell2017prototypical}. When the pre-trained CLIP is available as initialization, existing works seek to enhance cross-modal matching information via learning multi-modal prompts~\cite{wang2023attriclip,wang2022s}. MOE-Adapter~\cite{yu2024boosting} further extends the lightweight module selection process via mixture-of-experts~\cite{masoudnia2014mixture}, while PROOF~\cite{zhou2023learning} extends the representation ability of CLIP by appending new projection layers for new tasks. RAPF~\cite{huang2024class} learns the adapter modules for new tasks by decomposed parameter fusion to resist forgetting in the updating process.

 \section{Preliminaries}

 \subsection{Class-Incremental Learning}
 CIL is designed to continually build a unified classifier for all seen classes of the data stream~\cite{rebuffi2017icarl}. 
 We denote training sets as $\left\{\D^{1}, \D^{2}, \cdots, \D^{B}\right\}$, where each task $\D^{b}=\left\{\left(\x_{i}, y_{i}\right)\right\}_{i=1}^{n_b}$ contains $n_b$ instances.
 Each training instance $\x_i \in \R^D$  belongs to class $y_i \in Y_b$, \ie, the label space of task $b$ is $Y_b$. We have $Y_b  \cap Y_{b^\prime} = \varnothing$ for $b\neq b^\prime$. 
 This paper follows the {\bf exemplar-free} CIL setting~\cite{zhu2021prototype,wang2022learning,wang2022dualprompt}, where the model cannot hold any historical instances in the memory.
 In other words, we can only access data from $\D^b$ for model training when learning  the $b$-th incremental task.
 In CIL, the target is to build a unified classifier for all seen classes $\mathcal{Y}_b=Y_1 \cup \cdots Y_b$, \ie, find a model $f(\x): X\rightarrow\mathcal{Y}_b$ that minimizes the expected risk:
 \begin{equation} \label{eq:totalrisk} 
 	f^*=\underset{f \in \mathcal{H}}{\operatorname{argmin}} \; \mathbb{E}_{(\mathbf{x}, y) \sim \mathcal{D}_{t}^1\cup\cdots\mathcal{D}_{t}^b} \mathbb{I}\left(y \neq f(\mathbf{x})\right) \,.
 \end{equation}
 In Eq.~\ref{eq:totalrisk}, $\mathcal{H}$ represents the hypothesis space and $\mathbb{I}(\cdot)$ is the indicator function. $\mathcal{D}_{t}^b$ denotes the $b$-th task's data distribution. \looseness=-1 
 In this paper, we follow~\cite{zhou2023learning,yu2024boosting,huang2024class} and assume that a pre-trained CLIP model (contrastive language-image pre-training)~\cite{radford2021learning} is available as the initialization for $f(\x)$.
 Specifically, CLIP contains the visual and textual encoder, \ie, $g_i(\cdot): \R^D\rightarrow\R^d$, $g_t(\cdot): \R^{Dt}\rightarrow\R^d$, where the $D/Dt$-dimensional images and texts are projected to the same $d$-dimensional embedding space.
 During inference, it constructs the template text $\mathbf{t}_i$ using the class name, \ie, ``a photo of a [CLASS]$_i$'', and matches the query image to text embeddings of all classes:
 \begin{align} \label{eq:clip_pred}
	f_{y_i}(\x, \mathbf{t}_{i}) & =\frac{\exp \left(\cos \left(\z, \w_i  \right) / \tau\right)}{\sum_{j=1}^{|\mathcal{Y}_b|} \exp \left(\cos \left(\z, \w_j \right) / \tau\right)}
	\\
	\notag
	&= \frac{\exp \left(\cos \left(g_i(\x), g_t(\mathbf{t}_i)  \right) / \tau\right)}{\sum_{j=1}^{|\mathcal{Y}_b|} \exp \left(\cos \left(g_i(\x), g_t(\mathbf{t}_j) \right) / \tau\right)} \,,
\end{align}
 where $\cos(\cdot,\cdot)$ is cosine similarity and $\tau$ is temperature. $\z=g_i(\x)$ and $\w_i=g_t(\mathbf{t}_i)$ are visual and textual features, respectively. In Eq.~\ref{eq:clip_pred}, the logit is assigned by the relative similarity of the visual embedding to textual embedding.

 \subsection{Baselines in Class-Incremental Learning} \label{sec:baseline}

To overcome forgetting, there are two typical solutions for CIL with pre-trained models by learning prompts~\cite{jia2022visual,zhou2022learning}.

 \noindent\textbf{Learning Visual Prompts:} As fully-finetuning will harm the generalizability of the PTM, several works~\cite{wang2022learning,jia2022visual,wang2022dualprompt,smith2023coda} propose to freeze the pre-trained visual encoder and append visual prompts for subsequent tasks. 
 They design a prompt pool to select instance-specific prompts, with which the visual embedding is represented as:
 \begin{equation} \label{eq:l2p} 
 	 \z = \bar{g_i}\left(\x, \mathcal{P} \right)  \,,
 \end{equation}
 where $\bar{g_i}$ is the frozen image encoder, $\mathcal{P}$ is the prompt pool for prompt selection. By replacing $g_i(\x)$ in Eq.~\ref{eq:clip_pred} with Eq.~\ref{eq:l2p}, the model efficiently encodes downstream information into the visual prompts and tackles forgetting.
 
 \noindent\textbf{Learning Textual Prompts:} Since CLIP contains two branches for the visual and textual encoder, there are also works learning textual prompts instead of using the template text. Specifically, the textual input is formulated into a set of learnable prompts in CoOp~\cite{zhou2022learning}, \ie, $\tilde{\mathbf{t}}_i=$ [V]$_1$[V]$_2\cdots$[V]$_M$[CLASS]$_i$. In this way, the textual embedding is denoted as:
 \begin{equation} \label{eq:coop}
\w_i=\bar{g_t}\left( \tilde{\mathbf{t}}_i \right) \,,
 \end{equation}
 where $\bar{g_t}$ is the frozen text encoder. Similarly, we can replace the textual embedding in Eq.~\ref{eq:clip_pred} via Eq.~\ref{eq:coop} to encode task information into these learnable prompts.

 \noindent\textbf{Discussions:}  Eq.~\ref{eq:l2p} and Eq.~\ref{eq:coop} focus on different aspects to adjust the pre-trained CLIP into downstream tasks. However, when adapting the model with only visual prompts, the problem occurs since the model has to decompose the class name into fine-grained concepts to match the visual features. The sequential updating process will cause the degradation of such decomposition ability, resulting in the mismatch between visual and textual features.
  By contrast, if we utilize textual prompts, the model is only focused on the specific training instance and lacks holistic information about the specific class. When the testing data contains the image of the same class with different distributions, the model shall be confused due to the prompt-level overfitting. 
 Hence, the ideal text input should be general and holistic to match the visual features and convey all informative features that the target class possesses.

\section{{\scshape{Engine}}: External Knowledge Injection}

Noting that prompt-based methods are highly limited by the training instances, we need to extract comprehensive information for the model to adapt to downstream tasks while preventing forgetting. To encode {\em holistic information} into the model, we seek help from large language models like GPT-4~\cite{achiam2023gpt} to provide the general descriptions instead of using the template features. Apart from using detailed descriptions in the training process, we also make full use of them during inference by seeking pair-wise discriminative features. On the other hand, as external knowledge is continually injected into the model, we also design the injection unit expansion and prototype replay strategy. In the following sections, we first introduce how to inject external knowledge during model training and then introduce how to utilize it during inference. \looseness=-1

\begin{figure*}[t]
	\vspace{-6mm}
	\begin{center}
		{\includegraphics[width=2\columnwidth]{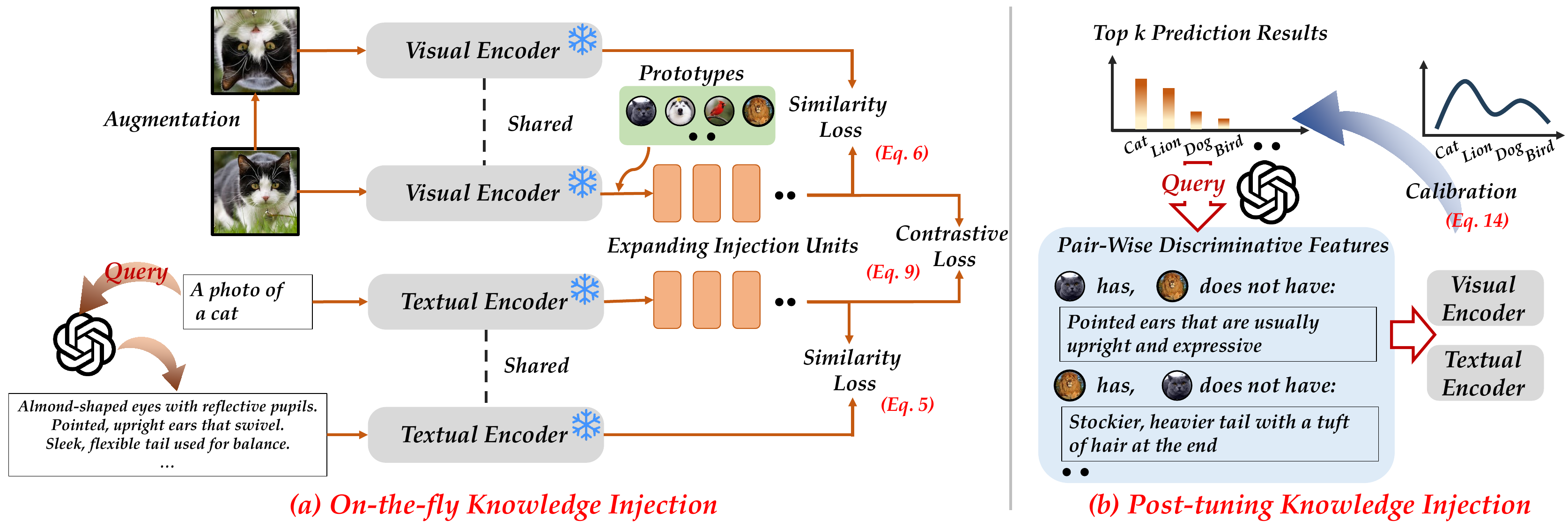}}
	\end{center}
	\vspace{-5mm}
	\caption{  Illustration of \mame. {\bf Left}: On-the-fly knowledge injection. We utilize GPT-4 to provide visual features of each class, and append injection units to encode external knowledge into the model. We also adopt random augmentation to enhance image diversity using the same network structure. To overcome forgetting, we append the prototypes of previous classes in each learning stage.  {\bf Right}: Post-tuning knowledge injection. We extract top-$k$ predictions and generate pair-wise discriminative features locally. The input image is forwarded with every textual feature to further refine the prediction.
	}
	\vspace{-7mm}
	\label{figure:teaser}
\end{figure*}

\subsection{On-the-fly Knowledge Injection}  \label{sec:trainng-injection}
Since the target is to encode general information into the CLIP during incremental learning, we first design the model training pipeline that can enhance recognition.
For example, a ``cat'' contains visual features like ``soft, short fur'', ``long, thin tail'', and ``round face with large eyes''. It would be better to match the cat image to these visual features than ``a photo of a cat'', since those features contain more discriminative and fine-grained information.
 Correspondingly, to enhance CLIP's knowledge of general class description, we seek help from GPT-4~\cite{achiam2023gpt} to provide {\em discriminative visual features}:
\begin{displayquote}
	 {\bf Q}: What are unique visual features of [CLASS]$_i$ in a photo? Focus on the key visual features.\\
	 {\bf A}: {\bf 1.} Long, thin tail that aids in balance. {\bf 2.}$\cdots$
\end{displayquote}
In this way, we can get a set of visual descriptions for each class, denoted as $\mathbf{d}_i$. A naive solution is to replace the template text $\mathbf{t}_i$ into $\mathbf{d}_i$ during training, while we propose a more intuitive way to inject the knowledge into the model. Specifically, we design a knowledge injection unit $u_t$ to dynamically encode these textual descriptors. We implement $u_t$ with a linear layer, \ie, $u_t(\cdot):\R^d\rightarrow\R^d$, and append it after the textual encoder. We have the similarity loss:
\begin{equation} \label{eq:proj_text_loss}
	\mathcal{L}_{t} = -\text{Sim} \left( u_t\left(\bar{g_t}\left( {\mathbf{t}}_i \right)\right), \bar{g_t}\left( {\mathbf{d}}_i \right) \right)  \,,
\end{equation}
where we use cosine similarity to calculate $\text{Sim}(\cdot,\cdot)$. In Eq.~\ref{eq:proj_text_loss}, we maximize the similarity between $u_t\left(\bar{g_t}\left( {\mathbf{t}}_i \right)\right)$ and $\bar{g_t}\left( {\mathbf{d}}_i \right)$, \ie, making the adapted textual feature similar to external descriptions. Since the textual encoder $\bar{g_t}$ is frozen, we can inject this external information into the injection unit $u_t$. In this way, the extracted textual features will contain detailed descriptions even using the template text, \eg, highlighting the features of ``long thin tail'' and ``round face'' when the input is ``a photo of a cat''. In the implementation, since GPT-4 will output a set of descriptions, we randomly choose one of the descriptions $\mathbf{d}_i$ to calculate Eq.~\ref{eq:proj_text_loss} in each iteration.\looseness=-1

\noindent\textbf{Learning external visual knowledge:} Similar to the textual branch, we can also inject visual information into the model during training. Given that we cannot fetch external data in the learning process, we utilize random data augmentation to adjust the input images, and denote the augmented input as $\mathcal{A}(\x)$. Correspondingly, we can learn the image knowledge injection unit $u_i(\cdot):\R^d\rightarrow\R^d$, and maximize the similarity between augmented and vanilla input: \looseness=-1
\begin{equation} \label{eq:proj_visual_loss}
	\mathcal{L}_{i}= -\text{Sim} \left( u_i\left(\bar{g_i}\left( \x \right)\right), \bar{g_i}\left( \mathcal{A}(\x) \right)\right)  \,.
\end{equation}
Similar to Eq.~\ref{eq:proj_text_loss}, Eq.~\ref{eq:proj_visual_loss} freezes the visual encoder $\bar{g_i}$, and only optimizes the image injection unit $u_i$. In the matching process, the model maximizes the similarity of the adapted visual feature to the augmented view, which helps the model extract diverse visual information.

\noindent\textbf{Learning task-specific injection unit:} In CIL, new tasks will emerge as data evolves, and sequentially tuning a single injection unit will result in forgetting previous knowledge. Consequently, we conduct the injection process for each incremental task to avoid forgetting. Specifically, we initialize a new injection unit $u_t^b$, $u_i^b$ when learning the $b$-th task, and freeze all previous injection units when learning the current task. With the set of visual and textual units, we utilize the summation of them for the final representation:
 \begin{equation} \label{eq:unit_aggregation-vis}
	\textstyle
	G_i(\x) =\sum_{p=1}^{b-1} \bar{u}_i^p \left(\bar{g_i}\left( \x \right) \right) + u_i^b \left(\bar{g_i}\left( \x \right) \right) 
\end{equation}
 \begin{equation} \label{eq:unit_aggregation-text}
	\textstyle
		G_t(\mathbf{t}_i) = \sum_{p=1}^{b-1} \bar{u}_t^p \left(\bar{g_t}\left( \mathbf{t}_i \right) \right) + u_t^b \left(\bar{g_t}\left( \mathbf{t}_i \right) \right)
\end{equation}
In the original CLIP, we have the cross-modal contrastive loss to match the image to the corresponding text description. Correspondingly, we utilize the injected visual and textual features to calculate the contrastive loss:
 \begin{align} \label{eq:unit-loss}
	\mathcal{L}_{c} &= \ell (f_\text{inj}(\x), y) \quad \text{where}\\ 
	\label{eq:unit-logit}
	f_{\text{inj},{y_i}}(\x) & =\frac{\exp \left(\cos \left(G_i(\x), G_t(\mathbf{t}_i)  \right) / \tau\right)}{\sum_{j=1}^{|\mathcal{Y}_b|} \exp \left(\cos \left(G_i(\x), G_t(\mathbf{t}_j) \right) / \tau\right)} \,.
\end{align}
In Eq.~\ref{eq:unit-loss}, we utilize the injected visual and textual features to calculate the contrastive loss, aiming to increase the similarity of pair-wise injected features.

\noindent\textbf{Preventing forgetting of injection unit} In the updating process, we combine Eq.~\ref{eq:proj_text_loss}, Eq.~\ref{eq:proj_visual_loss}, and Eq.~\ref{eq:unit-loss} to update the model. Since we cannot hold historical instances for replay, the model may still forget previous concepts due to the sequential updating. To this end, utilizing the substitution of previous concepts can create a calibration across all seen classes. Observing that the visual backbone $g_i$ is not updating throughout the learning process, we generate a subset of previous classes using visual prototypes. Specifically, during the learning process, we can calculate the average visual embedding of each class as:
\begin{equation}	\label{eq:prototype} \textstyle
	\p_k=\frac{{\sum_{j=1}^{|\mathcal{{D}}^b|}\mathbb{I}(y_j=k)g_i(\x_j)}}{{\sum_{j=1}^{|\mathcal{{D}}^b|}\mathbb{I}(y_j=k)}} \,,
\end{equation}
which stands for the most representative features of the corresponding class. We can treat them as the visual embedding of previous classes when learning new ones, \ie, we construct the auxiliary training set $\mathbf{P}=\{(\p_1,1),(\p_2,2),\cdots, (\p_{|\mathcal{Y}_b|},|\mathcal{Y}_b|)\}$, and optimize:
 \begin{equation} \label{eq:total-loss} \textstyle
	\min_{\{u_t^b, u_i^b\}} \sum_{(\x,y)\in \mathcal{D}^b \cup \mathbf{P}} \mathcal{L}_{t}+\mathcal{L}_{i}+ \mathcal{L}_{c} \,.
\end{equation}
Specifically, the prototype set $\mathbf{P}$ is only adopted in calculating $\mathcal{L}_c$. 
To enhance diversity across training stages, we also add Gaussian noise on the prototype embeddings, \ie, $\p^\prime=\p+\epsilon$, $\epsilon\in \mathcal{N}(0,\alpha^2\mathbf{I})$, and utilize $\p^\prime$ in Eq.~\ref{eq:total-loss}.
In this way, we can revisit previous knowledge to prevent forgetting when learning new injection units. Additionally, we also consider matching the visual features to these prototypes to further assist recognition~\cite{zhou2023learning,wanghard}.

\noindent\textbf{Summary of on-the-fly injection}: We visualize the training stage knowledge injection in Figure~\ref{figure:teaser} (left). \name learns two sets of knowledge injection units, aiming to encode the extra knowledge into the unit module. We utilize GPT-4 to extract informative class descriptors, and utilize similarity loss to encode them into the injection units. Similarly, data augmentation is utilized to provide visual information. Hence, the model can extract more informative features as it sees more tasks.
 To prevent forgetting, we also design the learning objective considering class prototypes. Since injection units are implemented with a single layer, the appended parameter size ($d\times d$) per unit is negligible for CLIP. Besides, since injection units are linear layers, they can be further re-parameterized by aggregating weights into a single one, \ie, $\sum_p u_i^p$ and $\sum_p u_t^p$.

\subsection{Post-tuning Knowledge Injection}

We have introduced the way to inject external knowledge during training in Section~\ref{sec:trainng-injection}. Apart from the model training stage, we can also utilize external knowledge to calibrate the prediction results during inference, and we call this step ``post-tuning knowledge injection''.  For example, after the model makes inference by matching visual and textual features, we can treat the current prediction results as a {\em preliminary result}, which needs further rectification. For the top-$k$ prediction results in this preliminary result, we treat them as {\em competitors} to the ground truth that could result in confusion. Hence, we need to construct a further prediction considering the image and all local competitors, which can focus more on the top-$k$ classes to refine the prediction. \looseness=-1

\noindent\textbf{Refining the results}: After learning the injection units, we can get the prediction results using the injected features for more comprehensive features. However, the model may still make wrong predictions, \eg, classifying a ``cat'' into a similar class ``lion'' since they have similar visual features. To refine the prediction results, we design an extra consolidation process using external knowledge. Denoting the prediction results of injected feature matching in Eq.~\ref{eq:unit-logit} as $f_\text{inj}(\x)$, we first extract the top-$k$ predictions as the competitor candidates, \ie, $\left\{f_{\text{inj}, i_1}(\x), f_{\text{inj}, i_2}(\x), \ldots, f_{\text{inj}, i_k}(\x)\right\}=\operatorname{Top}_k\left(f_{\text{inj}}(\x)\right)$, where $\operatorname{Top}_k\left(f_{\text{inj}}(\x)\right)$ indicates selecting the top-$k$ largest value in $f_\text{inj}(\x)$. Their labels are denoted as:
\begin{equation} \label{eq:confuse-labels}
	\left\{y_{i_1}, y_{i_2}, \ldots, y_{i_k}\right\}=\left\{y_i \mid f_{\text {inj }, i}(\mathbf{x}) \in \operatorname{Top}_k\left(f_{\text {inj }}(\mathbf{x})\right)\right\} \,.
\end{equation}
In Eq.~\ref{eq:confuse-labels}, the label set $\left\{y_{i_1}, y_{i_2}, \ldots, y_{i_k}\right\}$ corresponds to the possible confusing labels for the input image. To further refine the predictions, we aim for pair-wise disambiguation among this label set. We again utilize GPT-4 to generate {\em pair-wise} discriminative features:
\begin{displayquote}
	{\bf Q}: What are unique visual features of [CLASS]$_i$ compared to [CLASS]$_j$ in a photo? Focus on their key visual differences. \\
	{\bf A}: [CLASS]$_i$: {\bf 1.} Long, thin tail that aids in balance $\cdots$ [CLASS]$_j$: {\bf 1.} Stockier, heavier tail with a tuft of hair at the end $\cdots$
\end{displayquote}
With this process, GPT-4 provides the most discriminative features that can differentiate two classes, \eg, the differences in ``tail'' to differentiate ``cat'' from ``lion'' and vice versa. Given the top-$k$ classes, we can generate their pair-wise descriptors as $\mathbf{D}=\left[\mathbf{d}_{i j}\right]_{i, j=1}^k$, where $\mathbf{d}_{i j}$ denotes the descriptions to differ class $i$ from class $j$. We feed these fine-grained descriptions as the textual input, and match the query embedding to them:
\begin{equation} \label{eq:pt-logit} 
	f_{\text{pt},y_i}(\x)=\frac{1}{k-1}\sum_{j=1}^{j\neq i} f_{y_i}(\x,\mathbf{d}_{i j}) \,,
\end{equation}
where we utilize zero-shot CLIP in Eq.~\ref{eq:clip_pred} to match the query image to these pair-wise discriminative features, and average the similarities as the post-tuning logits. \looseness=-1

\noindent\textbf{Effect of post-tuning knowledge injection}: We visualize the post-tuning knowledge injection in  Figure~\ref{figure:teaser} (right). To discriminate similar concepts and refine the wrong predictions, we first extract the top-$k$ predictions of the model, and utilize GPT-4 to generate pair-wise discriminative features locally. Eq.~\ref{eq:pt-logit} provides the fine-grained local prediction among these top-$k$ classes, which can further help refine the results as data evolves. We adopt the zero-shot CLIP for Eq.~\ref{eq:pt-logit}, which does not require incremental updates.

\begin{table*}[t]
	\vspace{-7mm}
	\caption{Average and last performance comparison of different methods.  
		The best performance is shown in bold.  
		{\bf All methods are initialized with the same pre-trained CLIP without exemplars for a fair comparison.} }\label{tab:benchmark}
	\vspace{-3mm}
	\centering
	\resizebox{0.9\textwidth}{!}{%
		\begin{tabular}{@{}lccccccccccccccc}
			\toprule
			\multicolumn{1}{c}{\multirow{3}{*}{Method}}
			&
			\multicolumn{4}{c}{Aircraft }   & 
			\multicolumn{4}{c}{CIFAR100 }	&	
			\multicolumn{4}{c}{Cars }   
			\\ 
			& 
			\multicolumn{2}{c}{B0 Inc10}   & 
			\multicolumn{2}{c}{B50 Inc10}	&		
			\multicolumn{2}{c}{B0 Inc10}   & 
			\multicolumn{2}{c}{B50 Inc10}	& 
			\multicolumn{2}{c}{B0 Inc10}   & 
			\multicolumn{2}{c}{B50 Inc10}	& 
			\\  
			& 
			{$\bar{\mathcal{A}}$} & ${\mathcal{A}_B}$  
			& {$\bar{\mathcal{A}}$} & ${\mathcal{A}_B}$
			& {$\bar{\mathcal{A}}$} & ${\mathcal{A}_B}$ 
			&  {$\bar{\mathcal{A}}$} & ${\mathcal{A}_B}$  
			& {$\bar{\mathcal{A}}$} & ${\mathcal{A}_B}$
			& {$\bar{\mathcal{A}}$} & ${\mathcal{A}_B}$ 
			\\
			\midrule
			Finetune  &  3.16 & 0.96 & 1.72 & 1.05 & 7.84 & 4.44& 5.30 & 2.46& 3.14 & 1.10 & 1.54 & 1.13\\
			CoOp~\cite{zhou2022learning} & 14.54 & 7.14 & 13.05 & 7.77 & 47.00 & 24.24 & 41.23 & 24.12& 36.46 & 21.65& 37.40 & 20.87\\
			SimpleCIL~\cite{zhou2023revisiting} &59.24 & 48.09 & 53.05 & 48.09 & 84.15 & 76.63& 80.20 & 76.63& 92.04 & 86.85 & 88.96 & 86.85\\
			ZS-CLIP~\cite{radford2021learning} &26.66 & 17.22 & 21.70 & 17.22& 81.81 & 71.38& 76.49 & 71.38& 82.60 & 76.37& 78.32 & 76.37\\
			L2P~\cite{wang2022learning}  &47.19 & 28.29 &44.07&32.13& 82.74 & 73.03& 81.14 & 73.61& 76.63 & 61.82& 76.37 & 65.64 \\
			DualPrompt~\cite{wang2022dualprompt}  & 44.30& 25.83 &46.07&33.57 & 81.63 & 72.44& 80.12 & 72.57& 76.26 & 62.94& 76.88 & 67.55 \\
			CODA-Prompt~\cite{smith2023coda}  & 45.98 & 27.69 & 45.14 & 32.28& 82.43 & 73.43& 78.69 & 71.58& 80.21 & 66.47& 75.06 & 64.19 \\
			RAPF~\cite{huang2024class}   &  50.38  & 23.61 &  40.47 &  25.44 & 86.14 & 78.04 & 82.17 &  77.93  & 82.89 & 62.85 &  75.87 & 63.19\\
			\rowcolor{LightCyan}\name   & \bf69.69  &  \bf58.69 &  \bf64.38 &  \bf59.02 &  \bf86.92 & \bf79.22 & \bf83.15 &  \bf79.47 & \bf94.14 & \bf90.08 & \bf91.61 &  \bf90.03\\
		\end{tabular}
	}	
	\resizebox{0.9\textwidth}{!}{%
		\begin{tabular}{@{}lccccccccccccccc}
			\toprule
			\multicolumn{1}{c}{\multirow{3}{*}{Method}}
			& 
			\multicolumn{4}{c}{ImageNet-R }   & 
			\multicolumn{4}{c}{CUB }	&	\multicolumn{4}{c}{UCF }   
			\\ 
			& 
			\multicolumn{2}{c}{B0 Inc20}   & 
			\multicolumn{2}{c}{B100 Inc20}	&	\multicolumn{2}{c}{B0 Inc20}   & 
			\multicolumn{2}{c}{B100 Inc20}	& 
			\multicolumn{2}{c}{B0 Inc10}   & 
			\multicolumn{2}{c}{B50 Inc10}	& 
			\\  
			& 
			{$\bar{\mathcal{A}}$} & ${\mathcal{A}_B}$  
			& {$\bar{\mathcal{A}}$} & ${\mathcal{A}_B}$
			& {$\bar{\mathcal{A}}$} & ${\mathcal{A}_B}$ 
			&  {$\bar{\mathcal{A}}$} & ${\mathcal{A}_B}$  
			& {$\bar{\mathcal{A}}$} & ${\mathcal{A}_B}$
			& {$\bar{\mathcal{A}}$} & ${\mathcal{A}_B}$ 
			\\
			\midrule
			Finetune  & 1.37 & 0.43& 1.01 & 0.88& 2.06 & 0.64& 0.56 & 0.47& 4.51 & 1.59& 1.21 & 0.80\\
			CoOp~\cite{zhou2022learning} &60.73 & 37.52& 54.20 & 39.77& 27.61 & 8.57& 24.03 & 10.14& 47.85 & 33.46& 42.02 & 24.74\\
			SimpleCIL~\cite{zhou2023revisiting} & 81.06 & 74.48& 76.84 & 74.48& 83.81 & 77.52& 79.75 & 77.52& 90.44 & 85.68& 88.12 & 85.68\\
			ZS-CLIP~\cite{radford2021learning} &83.37 & 77.17& 79.57 & 77.17 & 74.38 & 63.06& 67.96 & 63.06& 75.50 & 67.64& 71.44 & 67.64\\
			L2P~\cite{wang2022learning}  &75.97 & 66.52 & 72.82 & 66.77&   70.87&57.93 & 75.64 &66.12 & 86.34 & 76.43& 83.95 & 76.62 \\
			DualPrompt~\cite{wang2022dualprompt}  &76.21 & 66.65 & 73.22 & 67.58&69.89 &57.46 & 74.40 &64.84 & 85.21 & 75.82& 84.31 & 76.35 \\
			CODA-Prompt~\cite{smith2023coda}  & 77.69 & 68.95 & 73.71 & 68.05& 73.12&62.98 &73.95&62.21 & 87.76 & 80.14&83.04 & 75.03 \\
			RAPF~\cite{huang2024class}  & 81.26  & 70.48 & 76.10 & 70.23 &  79.09 & 62.77& 72.82 & 62.93 & 92.28 & 80.33&90.31 & 81.55\\
			\rowcolor{LightCyan}\name    & \bf86.22 & \bf80.37  & \bf83.63 & \bf80.98 & \bf86.65 & \bf80.20& \bf82.59 & \bf79.30 & \bf94.35 & \bf90.03& \bf92.51 & \bf89.58\\
			
		\end{tabular}
	}
	
	\resizebox{0.9\textwidth}{!}{%
		\begin{tabular}{@{}lccccccccccccccc}
			\toprule
			\multicolumn{1}{c}{\multirow{3}{*}{Method}}
			&
			\multicolumn{4}{c}{SUN }   & 
			\multicolumn{4}{c}{Food }	&	\multicolumn{4}{c}{ObjectNet }   
			\\ 
			& 
			\multicolumn{2}{c}{B0 Inc30}   & 
			\multicolumn{2}{c}{B150 Inc30}	&		\multicolumn{2}{c}{B0 Inc10}   & 
			\multicolumn{2}{c}{B50 Inc10}	& 
			\multicolumn{2}{c}{B0 Inc20}   & 
			\multicolumn{2}{c}{B100 Inc20}	& 
			\\  
			& 
			{$\bar{\mathcal{A}}$} & ${\mathcal{A}_B}$  
			& {$\bar{\mathcal{A}}$} & ${\mathcal{A}_B}$
			& {$\bar{\mathcal{A}}$} & ${\mathcal{A}_B}$ 
			&  {$\bar{\mathcal{A}}$} & ${\mathcal{A}_B}$  
			& {$\bar{\mathcal{A}}$} & ${\mathcal{A}_B}$
			& {$\bar{\mathcal{A}}$} & ${\mathcal{A}_B}$ 
			\\
			\midrule
			Finetune  &4.51 & 1.59& 0.78 & 0.72& 3.49 & 1.71& 2.14 & 1.52& 1.34 & 0.47 & 0.69 & 0.54\\
			CoOp~\cite{zhou2022learning} &45.93 & 23.11 & 39.33 & 24.89& 36.01 & 14.18& 33.13 & 18.67& 21.24 & 6.29& 16.21 & 6.82\\
			SimpleCIL~\cite{zhou2023revisiting} & 82.13 & 75.58& 78.62 & 75.58& 87.89 & 81.65& 84.73 & 81.65& 52.06 & 40.13& 45.11 & 40.13\\
			ZS-CLIP~\cite{radford2021learning} &79.42 & 72.11& 74.95 & 72.11& 87.86 & 81.92& 84.75 & 81.92& 38.43 & 26.43 & 31.12 & 26.43\\	
			L2P~\cite{wang2022learning} &82.82 & 74.54 & 79.57 & 73.10 & 85.66 & 77.33& 80.42 & 73.13& 51.40 & 39.39& 48.91 & 42.83 \\
			DualPrompt~\cite{wang2022dualprompt} & 82.46 & 74.40 & 79.37 & 73.02& 84.92 &77.29& 80.00 & 72.75& 52.62 & 40.72& 49.08 & 42.92 \\
			CODA-Prompt~\cite{smith2023coda} &   83.34 & 75.71 & 80.38 & 74.17& 86.18 & 78.78& 80.98 & 74.13& 46.49 & 34.13& 40.57 & 34.13 \\
			RAPF~\cite{huang2024class}  & 82.13 & 72.47 & 78.04 & 73.10 & 88.57 & 81.15&\ 85.53 & 81.17&  48.67 & 27.43 & 39.28 &  28.73 \\
			\rowcolor{LightCyan}\name  & \bf85.04 & \bf78.54 & \bf81.57 & \bf78.45 & \bf89.81 & \bf83.89&\ \bf86.89 & \bf83.94&  \bf59.11  & \bf45.19  & \bf51.32 &  \bf44.99 \\
			\bottomrule
		\end{tabular}
	}
\end{table*}

\subsection{Summary of \name}
In \mame, we utilize external knowledge to help continual learning in two aspects, \ie, on-the-fly injection and post-tuning injection. With the help of GPT-4, we can extract more informative descriptors suitable for the incremental learning classes, and inject the external knowledge into the knowledge units. Besides, we also extract local descriptors to discriminate pair-wise classes in the post-tuning stage to help refine the predictions.
During training, we update the model via Eq.~\ref{eq:total-loss} to inject knowledge into the knowledge units.
 During inference, we aggregate the following outputs as the prediction:
\begin{equation} \label{eq:pred}  
	f(\x)=f_\text{inj}(\x)+f_\text{pt}(\x) \,.
\end{equation}

\section{Experiments}

In this section, we conduct extensive experiments on nine benchmark datasets and compare \name to state-of-the-art methods. We visualize the incremental learning curve and provide ablation studies on prompt engineering and parameter sensitivity to test the model's robustness. Besides, we also visualize the learned embeddings and logit changes after injecting external knowledge to show its effectiveness. We provide more results in the supplementary.

\begin{table}[t]
	\vspace{-3mm}
	\caption{ Comparison to traditional exemplar-based CIL methods. \name does not use any exemplars. 
	}  
	\label{tab:benchmark-typicalmethods}
	\centering
	\vspace{-3mm}
	\resizebox{0.98\columnwidth}{!}{%
		\begin{tabular}{@{}lccccccccc }
			\toprule
			\multicolumn{1}{l}{\multirow{2}{*}{Method}} &
			\multicolumn{1}{l}{\multirow{2}{*}{Exemplars}} & 
			\multicolumn{2}{c}{Aircraft B0 Inc10} & \multicolumn{2}{c}{SUN B0 Inc30}  \\
			& & {$\bar{\mathcal{A}}$} & ${\mathcal{A}_B}$  
			& {$\bar{\mathcal{A}}$} & ${\mathcal{A}_B}$
			\\
			\midrule
			iCaRL~\cite{rebuffi2017icarl}& 20 / class & 53.60  & 43.98 & 78.56 & 67.30 \\
			MEMO~\cite{zhou2022model} &20 / class &42.24 & 25.41 & 81.48 & 73.45\\
			PROOF~\cite{zhou2023learning} & 20 / class & 61.00 & 53.59  & 83.57 & 77.28\\
			\rowcolor{LightCyan}	\name  &\bf 0 & \bf 69.69  & \bf58.69  & \bf 85.04 & \bf 78.54\\
			\bottomrule
		\end{tabular}
	}
	\vspace{-3mm}
\end{table}

\begin{figure*}
	\vspace{-7mm}
	\centering
	\begin{subfigure}{0.33\linewidth}
		\includegraphics[width=1\columnwidth]{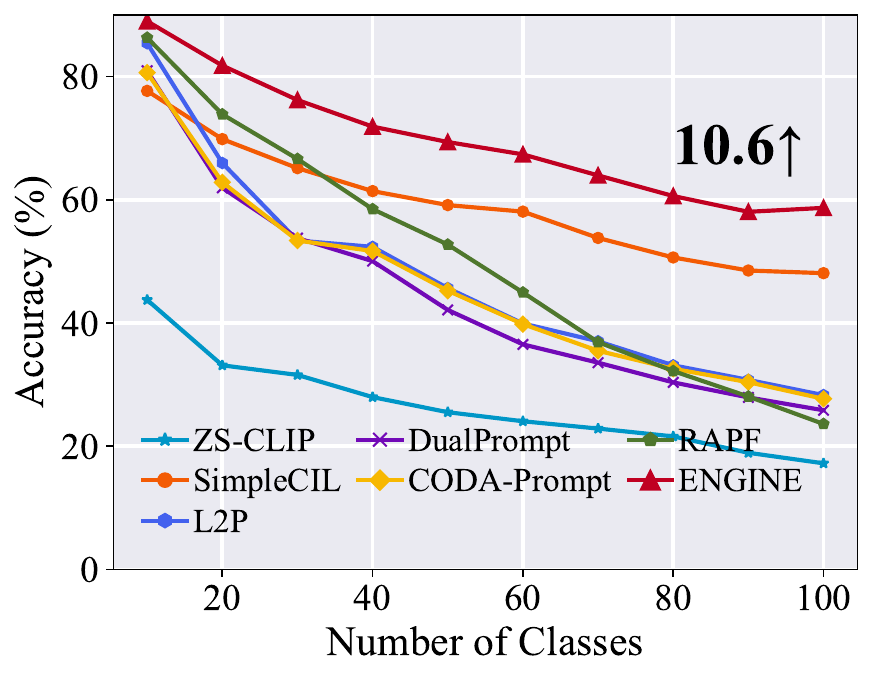}
		\caption{Aircraft Base0 Inc10}
		\label{fig:benchmark-aircraft}
	\end{subfigure}
	\hfill
	\begin{subfigure}{0.33\linewidth}
		\includegraphics[width=1\linewidth]{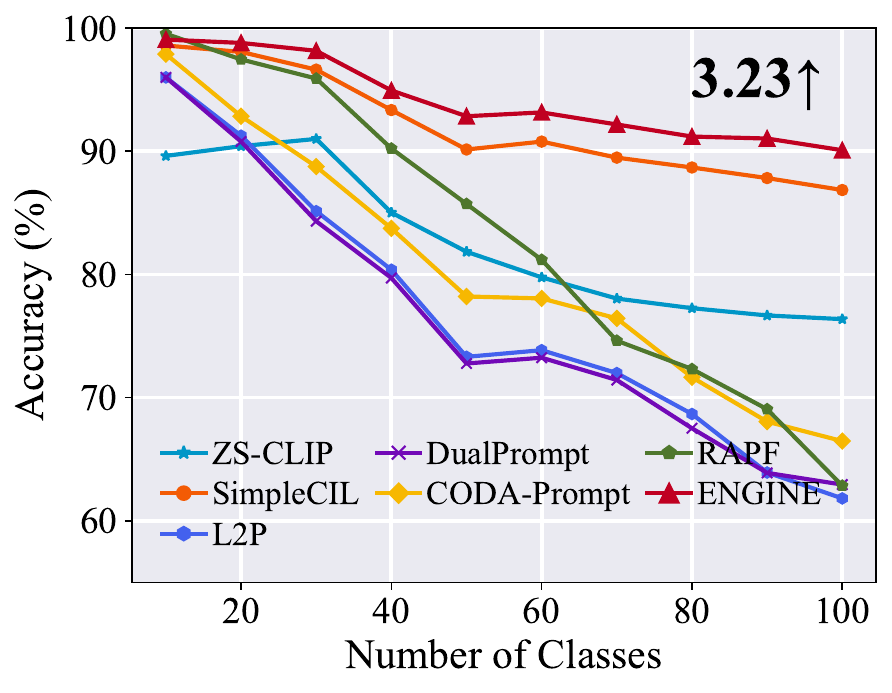}
		\caption{Cars Base0 Inc10}
		\label{fig:benchmark-cars}
	\end{subfigure}
	\hfill
	\begin{subfigure}{0.33\linewidth}
		\includegraphics[width=1\columnwidth]{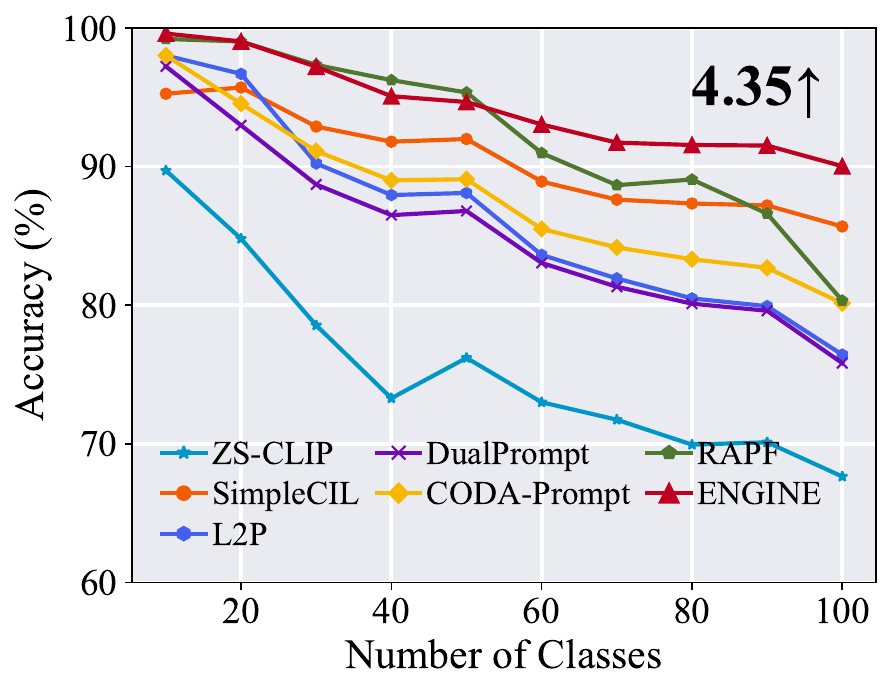}
		\caption{UCF Base0 Inc10}
		\label{fig:benchmark-ucf}
	\end{subfigure}
	\vspace{-7mm}
	\caption{\small 	Incremental performance of different methods. We report the performance gap after the last incremental stage of \name and the runner-up method at the end of the line.    All methods utilize the same CLIP pre-trained weight. More figures are shown in supplementary.}
	\vspace{-3mm}
	\label{fig:benchmark}
\end{figure*}

\subsection{Implementation Details}
\noindent {\bf Dataset}: We follow~\cite{zhou2023learning,zhou2022learning,wang2022learning} to evaluate the performance on nine benchmark datasets that have domain gap to CLIP's pre-training dataset, \ie, {CIFAR100}~\cite{krizhevsky2009learning}, {CUB200}~\cite{WahCUB2002011}, {ObjectNet}~\cite{barbu2019objectnet}, {ImageNet-R}~\cite{hendrycks2021many}, {FGVCAircraft}~\cite{maji2013fine}, {StanfordCars}~\cite{krause20133d}, {Food101}~\cite{bossard2014food}, {SUN397}~\cite{xiao2010sun} and {UCF101}~\cite{soomro2012ucf101}. 
Following~\cite{zhou2023learning}, we utilize the sampled 100 classes from CIFAR100, Aircraft, Cars, Food, UCF, 200 classes from CUB200, ObjectNet, ImageNet-R, and 300 classes from SUN to ease the data split.
 More details are reported in the supplementary.

\noindent {\bf Dataset split:} Following~\cite{rebuffi2017icarl,wang2022learning}, we use `B-$m$ Inc-$n$' to split the classes in CIL.
$m$ indicates the number of classes in the first stage, and $n$ represents that of every following stage. 
We follow~\cite{rebuffi2017icarl} to randomly shuffle the class order with random seed 1993 for all compared methods, and keep this same for every method.

\begin{table}[t]
	\vspace{-3mm}
	\caption{ Results when all methods use the same textual description generated by GPT-4 and data augmentation.
	}  
	\label{tab:benchmark-prompt}
	\centering
	\vspace{-3mm}
	\resizebox{0.98\columnwidth}{!}{%
		\begin{tabular}{@{}lccccccccc }
			\toprule
			\multicolumn{1}{l}{\multirow{2}{*}{Method}} &
			\multicolumn{1}{c}{\multirow{2}{*}{Template}} & 
			\multicolumn{2}{c}{ImageNet-R B0 Inc20} & \multicolumn{2}{c}{CIFAR B0 Inc10}  \\
			& & {$\bar{\mathcal{A}}$} & ${\mathcal{A}_B}$  
			& {$\bar{\mathcal{A}}$} & ${\mathcal{A}_B}$
			\\
			\midrule
			ZS-CLIP& GPT Generated & 83.68 &77.59 & 82.26& 71.70\\
			RAPF& GPT Generated & 81.89&71.04 & 86.49& 78.52 \\
			\rowcolor{LightCyan}	\name  & GPT Generated & \bf 86.22& \bf 80.37  & \bf 86.92 & \bf 79.22\\
			\bottomrule
		\end{tabular}
	}
	\vspace{-5mm}
\end{table}

\noindent {\bf Comparison methods:} We first compare to SOTA pre-trained model-based CIL algorithms, \eg, L2P~\cite{wang2022learning}, DualPrompt~\cite{wang2022dualprompt}, CODA-Prompt~\cite{smith2023coda}, SimpleCIL~\cite{zhou2023revisiting}. Besides, we also compare to SOTA CLIP-based CIL algorithms, \eg, CoOp~\cite{zhou2022learning}, PROOF~\cite{zhou2023learning}, RAPF~\cite{huang2024class}. 
The baseline that finetunes CLIP for incremental tasks is denoted as Finetune. {\bf All methods are deployed with the same CLIP as initialization.}

\noindent {\bf Training details:}
The experiments are deployed with NVIDIA 4090 using PyTorch~\cite{paszke2019pytorch}. We follow~\cite{zhou2023learning,huang2024class} to consider CLIP with ViT-B/16 for all compared methods for {\em fair comparison}.
For vision-based methods that cannot utilize the textual prompt (\eg, L2P, DualPrompt, CODA-Prompt), we utilize CLIP's visual branch as their initialization.
 We report the results using LAION-400M pre-trained CLIP~\cite{ilharco_gabriel_2021_5143773} in the main paper, and report results using OpenAI~\cite{radford2021learning} in the supplementary.
In \mame, we use SGD optimizer with a batch size of $64$ to optimize the model for $10$ epochs. The learning rate decays from $0.05$ with cosine annealing. We set the prototype noise ratio $\alpha$ to 0.25 and $k$ in post-tuning to 5. We use OpenAI GPT-4o mini~\cite{achiam2023gpt} to generate the prompts for textual knowledge and AutoAugment~\cite{cubuk2019autoaugment} to augment the visual features. We explore other LLMs to extract external knowledge in the supplementary.
The source code will be publicly available upon acceptance.

\noindent {\bf Evaluation metric:} Following~\cite{rebuffi2017icarl,zhou2023learning}, we use $\mathcal{A}_b$ to represent the model's accuracy after the $b$-th stage. Specifically, we adopt $\mathcal{A}_B$ (the performance after the last stage) and $\bar{\mathcal{A}}=\frac{1}{B}\sum_{b=1}^{B}\mathcal{A}_b$ (average performance along incremental stages) as measurements.

\begin{figure}
	\vspace{-3mm}
	\centering
	\begin{subfigure}{0.49\linewidth}
		\includegraphics[width=1\columnwidth]{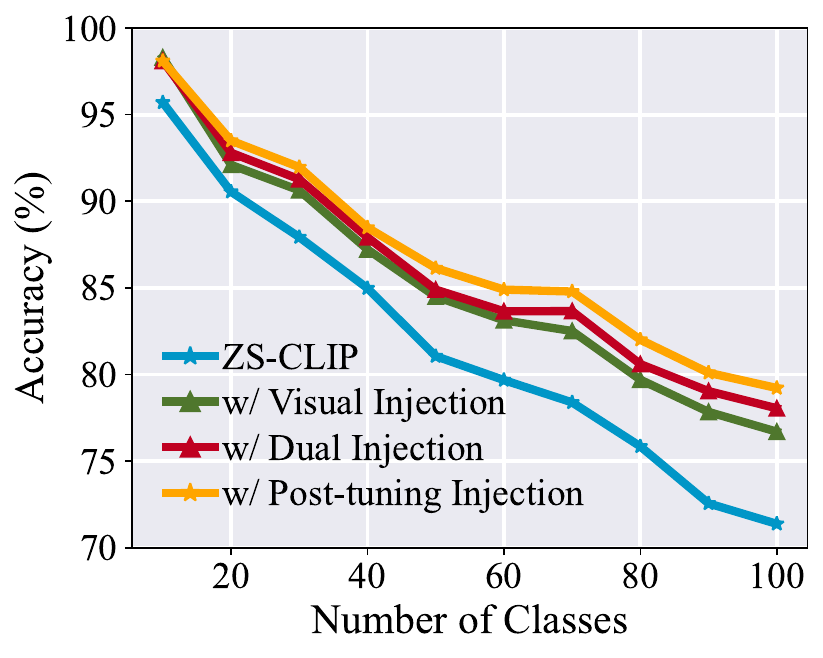}
		\caption{Ablation study}
		\label{fig:ablation}
	\end{subfigure}
	\hfill
	\begin{subfigure}{0.49\linewidth}
		\includegraphics[width=1\linewidth]{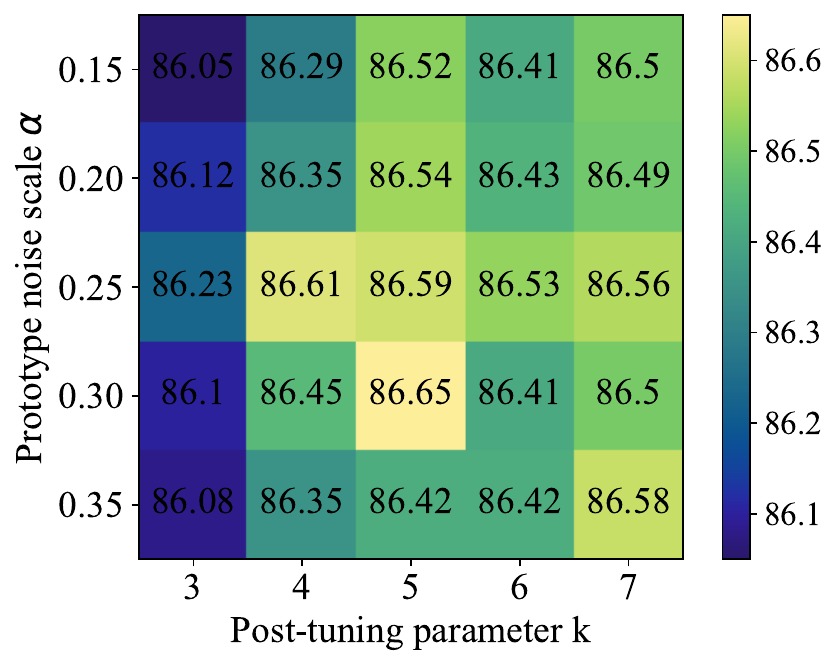}
		\caption{Parameter sensitivity}
		\label{fig:sensitivity}
	\end{subfigure}
	\vspace{-3mm}
	\caption{\small Ablation study and parameter sensitivity.}
	\vspace{-4mm}
	\label{fig:ablation-and-sensitivity}
\end{figure}

\subsection{Benchmark Comparison}

We first compare \name to other state-of-the-art methods on the benchmark datasets and report the results in Table~\ref{tab:benchmark} and Figure~\ref{fig:benchmark}. Specifically, we find \name outperforms current SOTA by 3\%$\sim$10\% on these benchmark datasets.
 We find the baseline method Finetune has the worst performance, indicating that the decomposition ability to tackle the class template into fine-grained features is totally forgotten.
 For visual prompt-based methods like L2P, DualPrompt, and CODA-Prompt, the performance is limited due to the inability to utilize textual information. For the textual prompt tuning method CoOp, we find that it achieves poor performance due to the forgetting of textual prompts.
For other CLIP-based methods like RAPF, our method still shows substantial improvement, verifying its strong continual learning ability without forgetting. \looseness=-1

Apart from these non-exemplar-based methods, we also consider typical exemplar-based CIL methods for comparison. We report the comparison results in Table~\ref{tab:benchmark-typicalmethods}, where all methods are initialized with the same CLIP. As we can infer from the table, \name still outperforms them even without using exemplars.

Finally, what if all CLIP-based methods utilize the same external knowledge? Correspondingly, we also consider using the same generated prompt and data augmentation for ZS-CLIP and RAPF, and report the results in Table~\ref{tab:benchmark-prompt}. In this table, we replace the template text of all methods using the same text in Section~\ref{sec:trainng-injection} and adopt the same data augmentation for image input. 
We find \name still outperforms them even using the same prompt, indicating that \name can better utilize the external information for incremental learning.

\subsection{Further Analysis}

\noindent\textbf{Ablation study}:  We first conduct ablations to investigate the effectiveness of each component in  \mame. We report the incremental performance on CIFAR100 B0 Inc10 in Figure~\ref{fig:ablation}. We can infer that `{\bf ZS-CLIP}' has the worst performance since the downstream distribution is different from the pre-training stage, and we view it as the baseline. Correspondingly, we first equip the model with visual knowledge injection (Eq.~\ref{eq:proj_visual_loss}), and denote the model as `{\bf w/ Visual Injection}'. It shows that the performance drastically improves as the model has been adapted to extract more informative visual features. Similarly, we then append the textual information using Eq.~\ref{eq:proj_text_loss}, denoted as `{\bf w/ Dual Injection}.' As shown in the figure, using external knowledge for both modalities can further improve the performance than visual knowledge only. Finally, we equip the model with post-tuning knowledge injection, and utilize Eq.~\ref{eq:pred} for inference (\ie, `{\bf w/ Post-tuning Injection}'). This corresponds to the full version of \mame, and we find it achieves the best performance. Ablations verify that every component in \name boosts the CIL performance.

\begin{figure}
	\vspace{-5mm}
	\centering
	\begin{subfigure}{0.49\linewidth}
		\includegraphics[width=1\columnwidth]{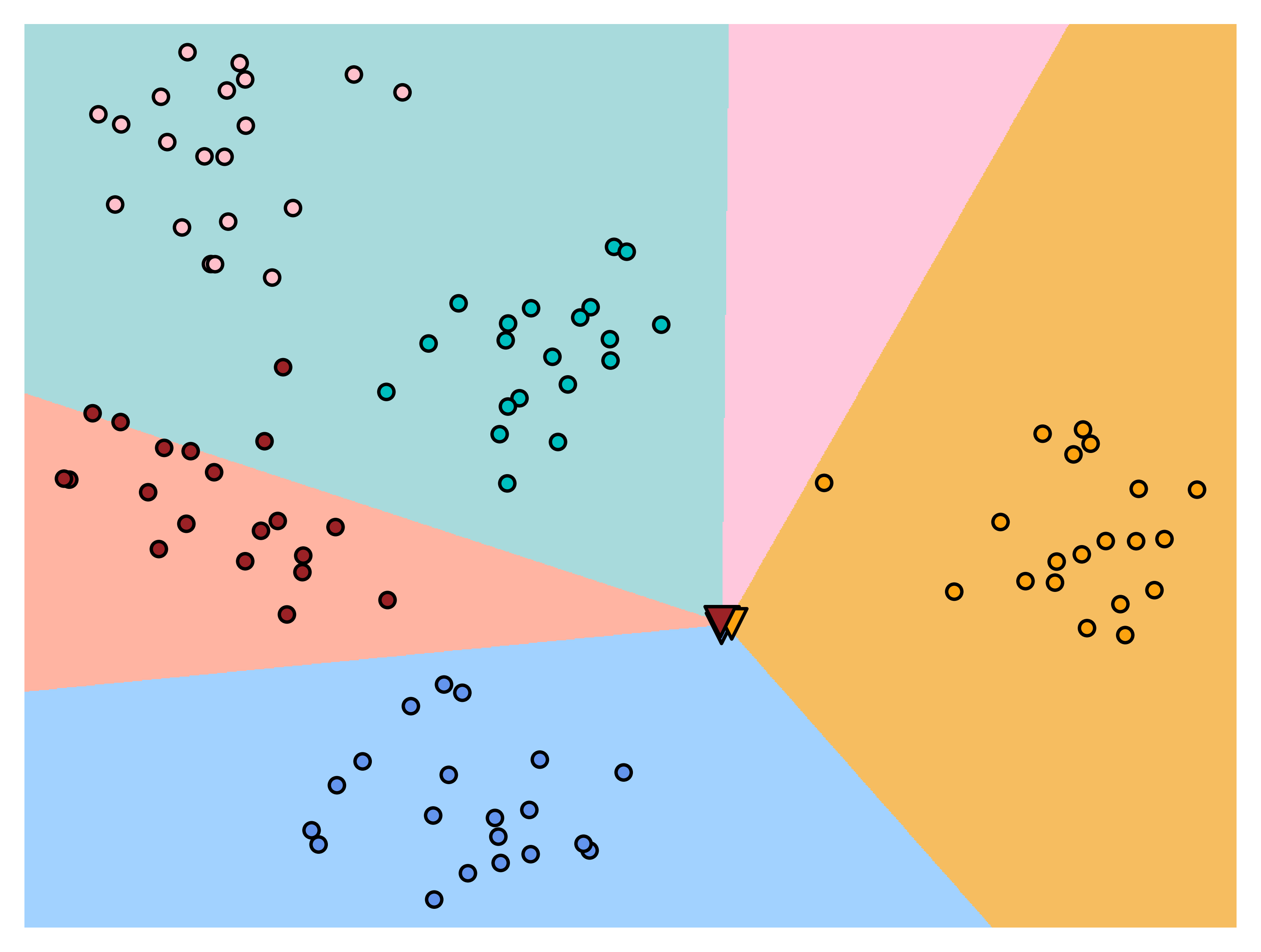}
		\caption{ZS-CLIP, single stage}
		\label{fig:vis-a}
	\end{subfigure}
	\hfill
	\begin{subfigure}{0.49\linewidth}
		\includegraphics[width=1\linewidth]{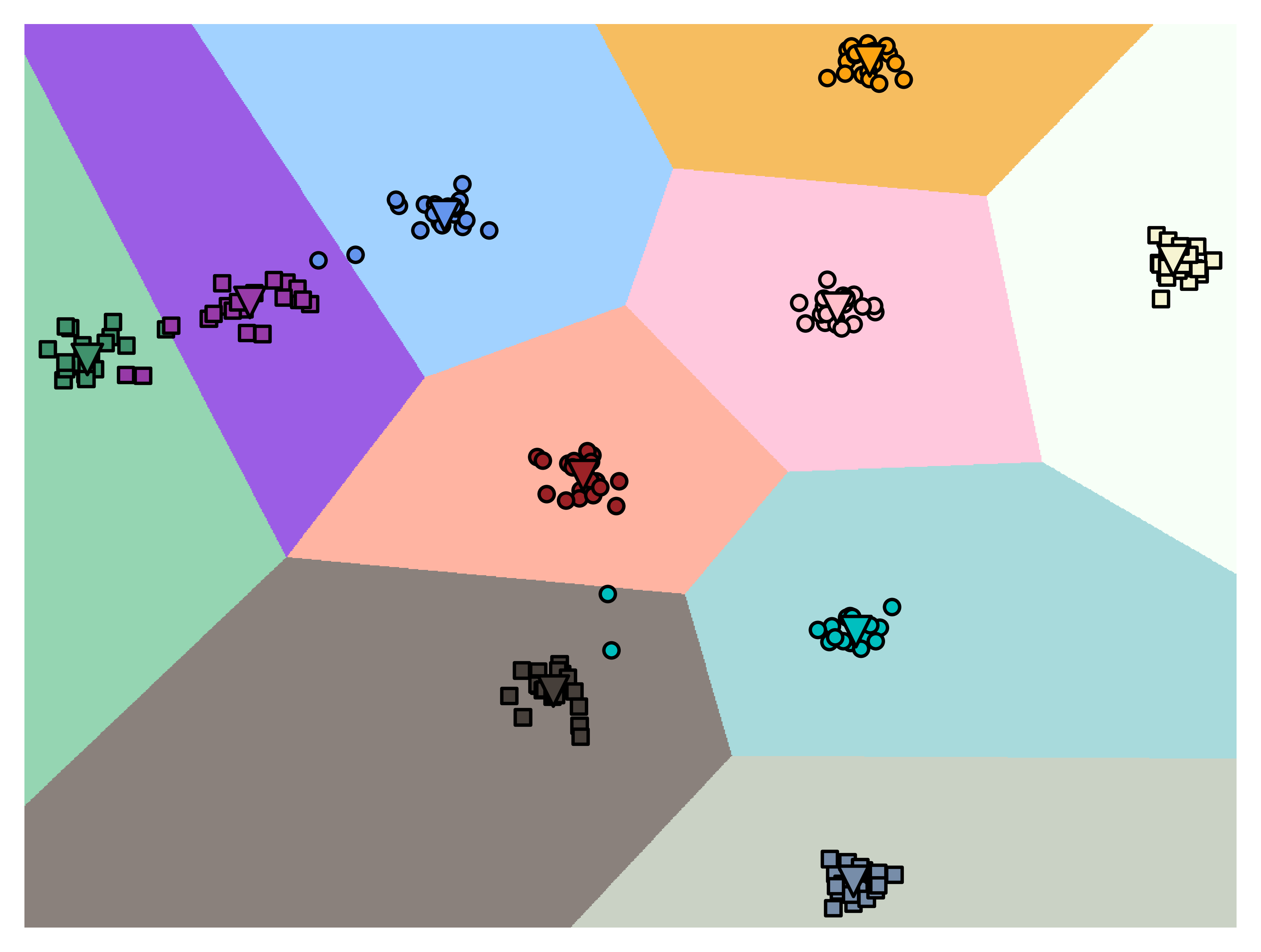}
		\caption{\mame, two stages}
		\label{fig:vis-b}
	\end{subfigure}
	\vspace{-3mm}
	\caption{\small t-SNE~\cite{van2008visualizing} visualizations of zero-shot CLIP and \mame. \name aligns the cross-modal features without forgetting. We represent the visual features of the first task with dots and classes of the second task with squares. The textual features are represented by triangles, and we utilize the shadow area to denote the decision boundary.}
	\vspace{-4mm}
	\label{fig:vis}
\end{figure}

\noindent\textbf{Parameter robustness}: There are two hyperparameters in \mame, \ie, the prototype noise scale $\alpha$ in Section~\ref{sec:trainng-injection} and post-tuning candidate $k$ in Eq.~\ref{eq:pt-logit}. We conduct experiments on CUB B0 Inc20 to investigate the robustness by changing these parameters. Specifically, we choose $\alpha$ among $\{0.15, 0.2, 0.25, 0.3, 0.35\}$, and $k$ among $\{3, 4, 5, 6, 7\}$. We report the last performance in Figure~\ref{fig:sensitivity}. As shown in the figure, the performance is robust with the change of these parameters, and we suggest $\alpha=0.25, k=5$ as default.

\noindent\textbf{Visualizations}: In this section, we utilize t-SNE~\cite{van2008visualizing} to visualize the cross-modal features learned by \name on CIFAR100 B0 Inc5. 
We represent the visual features of the first task with dots ($\bigcirc$) and classes of the second task with squares ($\square$). The textual features are represented by triangles ($\triangle$), and we utilize the shadow area to denote the decision boundary divided by the textual features.
Specifically, we first visualize the instances of the first stage in zero-shot CLIP's embedding space in Figure~\ref{fig:vis-a}. The results indicate that pre-trained CLIP has difficulty in aligning the visual and textual features since they are situated in two distinct areas in the embedding space~\cite{liang2022mind}. While for Figure~\ref{fig:vis-b}, we visualize the embedding of \name across two incremental stages. As we can infer from the figure, \name has two advantages: {\bf 1)} The visual and textual features are aligned in the same cluster, and {\bf 2)} The model has not forgotten previous classes when learning new ones.

\begin{figure}
	\vspace{-5mm}
	\centering
	{\includegraphics[width=.99\columnwidth]{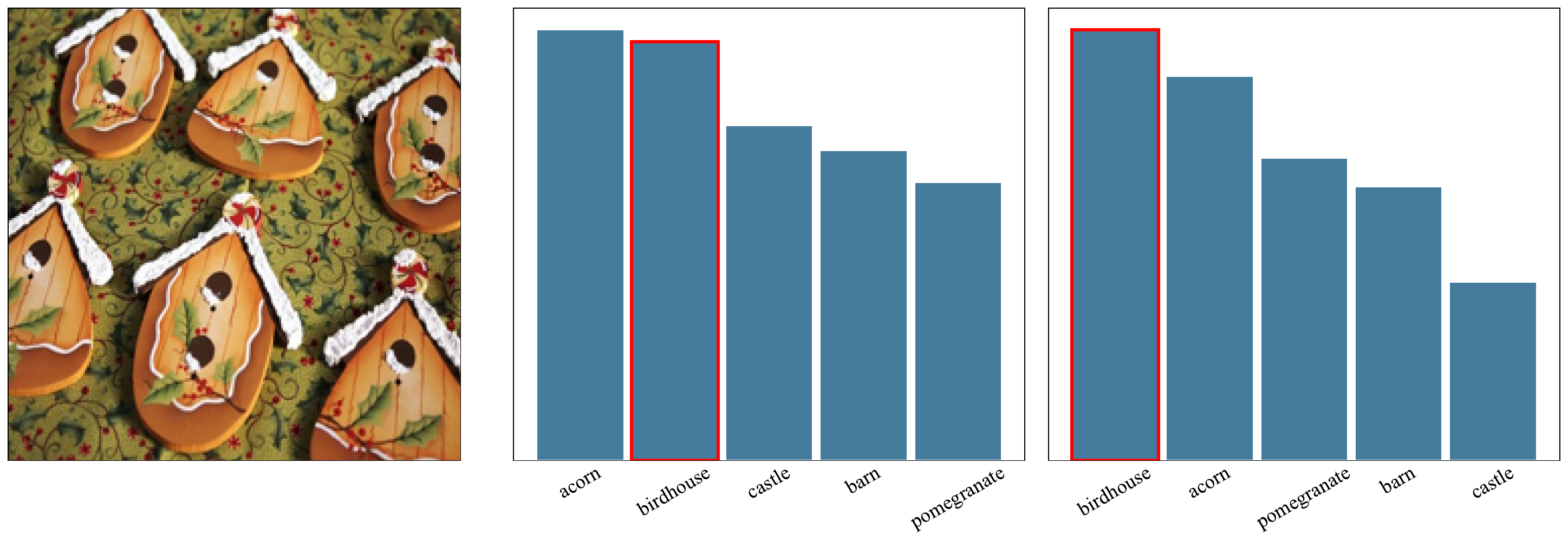}}
	{\includegraphics[width=.99\columnwidth]{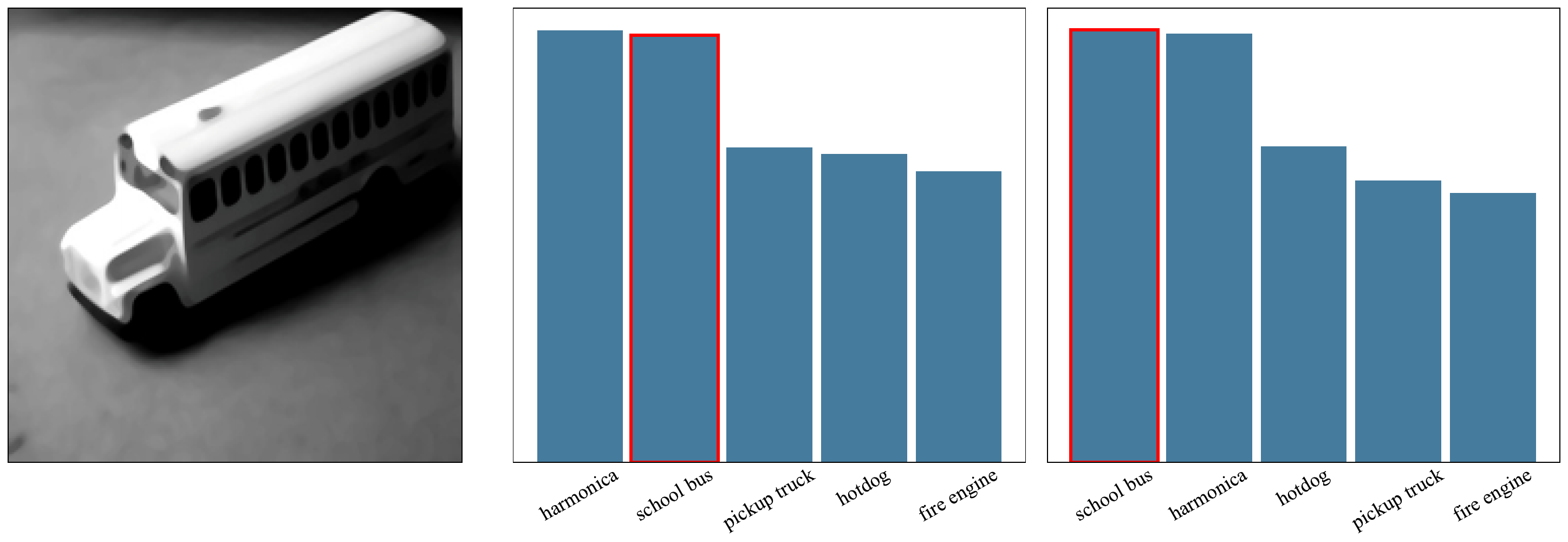}}
	\vspace{-4mm}
	\caption{\small {\bf Left:} Input. {\bf Middle:} Top-5 predictions before post-tuning knowledge injection.  {\bf Right:} Top-5 predictions after post-tuning. More cases are shown in the supplementary.}
	\vspace{-3mm}
	\label{fig:further-analysis}
\end{figure}

We also visualize the model output before and after post-tuning knowledge injection on ImageNet-R in Figure~\ref{fig:further-analysis}. We utilize bars with red edges to denote the ground-truth class. As we can infer from the figure, post-tuning knowledge injection can help the model refine the predictions by generating local comparisons. For the case on top of Figure~\ref{fig:further-analysis}, the model can focus on the differences between ``acorn'' and ``birdhouse'' to make the final prediction. 
 
\section{Conclusion} \label{sec:conclusion}

Class-incremental learning is essential to real-world applications. This paper aims to enable CLIP with CIL ability by injecting external knowledge. Specifically, we consider on-the-fly knowledge injection by learning dual-branch injection units. We augment images for the visual branch and generate detailed feature descriptions with GPT-4 for the textual branch. By maximizing the similarity between external information and the injection module, we encode informative features into the pre-trained CLIP. Besides, we also consider post-tuning knowledge injection with local class information to calibrate the predictions. Extensive experiments verify \mame's effectiveness.
\\\noindent\textbf{Limitations and future works:} This paper relies on GPT-4 to provide expert knowledge for specific tasks, which may be impossible for the specific cases where GPT-4 cannot generalize. Future works include designing other formats of external knowledge, \eg, knowledge graphs.

\section*{Acknowledgments}
This work is partially supported by NSFC (U23A20382, 62476123, 62376118, 62250069), Fundamental Research Funds for the Central Universities (2024300373,14380021), CCF-Tencent Rhino-Bird Open Research Fund (RAGR20240101), Postdoctoral Fellowship Program of CPSF (GZB20250389), China Postdoctoral Science Foundation (2025M771535), and Collaborative Innovation Center of Novel Software Technology and Industrialization.

{
    \small
    \bibliographystyle{ieeenat_fullname}
    \bibliography{paper}
}

\begin{center}
	\textbf{\large Appendix }
\end{center}
\appendix

\setcounter{section}{0}
\renewcommand{\thesection}{\Roman{section}}
\makeatletter

	In the main paper, we propose ExterNal knowledGe INjEction (\mame)
	for CLIP-based CIL. To enhance knowledge transfer from outside the dataset, we propose a dual-branch injection tuning framework that encodes informative knowledge from both visual and textual modalities. The visual branch is enhanced with data augmentation to enrich the visual features, while the textual branch leverages GPT-4 to rewrite discriminative descriptors. In addition to this on-the-fly knowledge injection, we also implement post-tuning knowledge by re-ranking the prediction results during inference. With the injected knowledge, the model can better capture informative features for downstream tasks as data evolves. 
	
	In this supplementary material, we provide more details about \mame, including more implementation details and experimental results.
	\begin{itemize}
		 \item Section~\ref{sec:supp_further_ablation}  introduces further analysis of \mame, including multiple runs, running time comparison, trainable parameter analysis, results of different backbones and different LLMs, and other baselines.
		 \item Section~\ref{sec:supp_intro} introduces the details of compared methods.
		 \item Section~\ref{sec:supp_full_results} provides supplementary results of benchmark datasets to the main paper.
		 \item Section~\ref{sec:supp_vis} provides more visualizations, including the visualization of prediction results and generated textual descriptions.
	\end{itemize}

\section{More results} \label{sec:supp_further_ablation}

This section includes more results on \mame, including the results with multiple runs using different random seeds, the details about trainable parameters, running time comparison, results with different pre-trained weights, and results with different LLMs to generate textual descriptions.

\begin{figure}
	\centering
	\includegraphics[width=0.8\columnwidth]{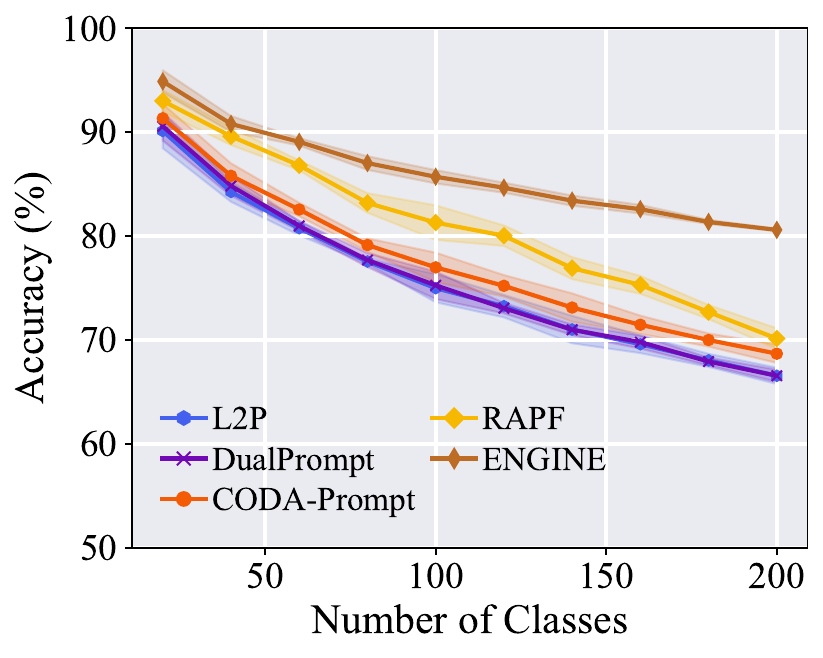}
	\caption{Results on ImageNet-R B0 Inc20 with multiple runs.  {\bf \name consistently outperforms other methods by a substantial margin. }}
	\label{fig:supp-multiple}
\end{figure}

\subsection{Multiple Runs}

In the main paper, the experimental results are conducted via splitting the classes with random seed 1993, which is a common practice in CIL~\cite{rebuffi2017icarl}. To investigate the robustness of different methods, we also consider running the experiments multiple times using different random seeds. Specifically, we conduct the class split using random seed \{1993,1994,1995,1996,1997\}, and calculate the average performance and standard variance. We report the results in Figure~\ref{fig:supp-multiple} on ImageNet-R B0 Inc20. 

As shown in the figure, \name shows more robust results against other baselines. We find the results of \name consistently outperform other competitors in multiple runs.

\subsection{Trainable parameters}

In this paper, we design \name by extending knowledge injection unit per task. During inference, the injected features are aggregated as the final feature. As illustrated in Section 4.1 of the main paper, we can reparameterize these injection units by adding the weights since they are linear layers, \ie, $\sum_p u_i^p$ and $\sum_p u_t^p$. Hence, the extra parameter size can be squeezed from $2\times b\times d \times d$ to  $2\times d \times d$. We further report the number of trainable parameters in each compared method in Table~\ref{tab:supp-trainable-params}. As we can infer from the table, \name has the same scale of trainable parameters compared to other competitors, while having the best performance.

	\begin{table}[t]
	\caption{ Number of trainable parameters on CIFAR100 B0 Inc10 setting. }
	\label{tab:supp-trainable-params}
	\centering
	\begin{tabular}{@{}lcccccc}
		\toprule
		Method & Trainable Parameters \\
		\midrule
		L2P &  161330 \\
		DualPrompt & 333412    \\
		CODA-Prompt& 3916900 \\
		RAPF& 262144\\
		\name   & 524288   \\
		\bottomrule
	\end{tabular}
\end{table}

\subsection{Running Time Comparison}

\begin{figure}[h]
	\centering
		\includegraphics[width=0.9\columnwidth]{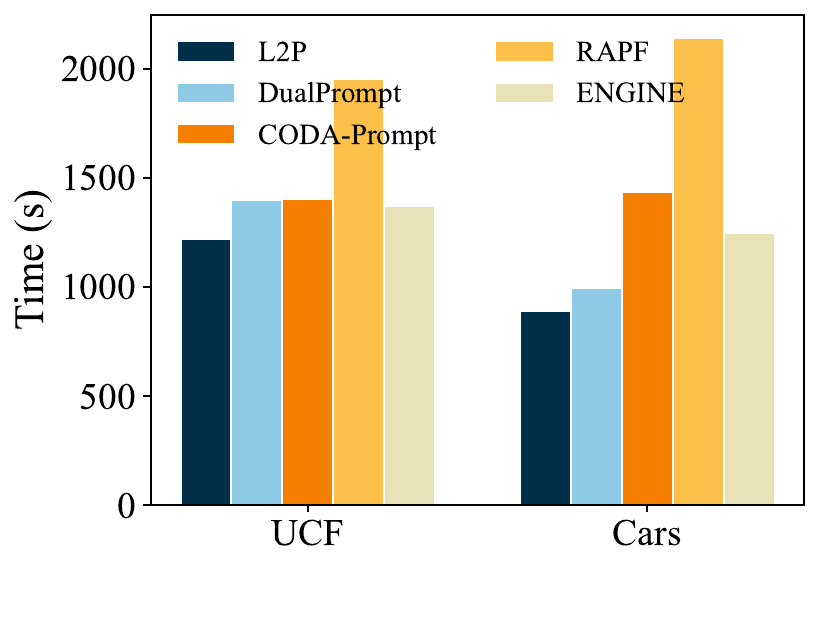}
		\caption{Running time comparison. {\bf \name has similar training time to other compared methods while having the best performance.}}
	\label{fig:supp-running-time}
\end{figure}

In this section, we report the running time comparison of different methods. We utilize a single NVIDIA 4090 GPU to run the experiments and report the results in Figure~\ref{fig:supp-running-time}. As we can infer from the figure, \name requires less running time than CODA-Prompt and RAPF, while having the best performance.
Experimental results verify the effectiveness of \mame.

Besides, the post-tuning process can be directly sped up with simple modifications.
Specifically, we pre-compute the one-to-one text pairs and generate all corresponding text embeddings in advance. Assuming there are $100$ classes and each pair contains $n$ distinct descriptions, the total number of embeddings amounts to approximately $99 \times  100 \times n$. As a result, we only need to compute the descriptions in advance and subsequently retrieve the pre-calculated embeddings during inference, significantly reducing the computational overhead.

\subsection{Different backbones}
\begin{figure}
	\centering
	\includegraphics[width=0.8\columnwidth]{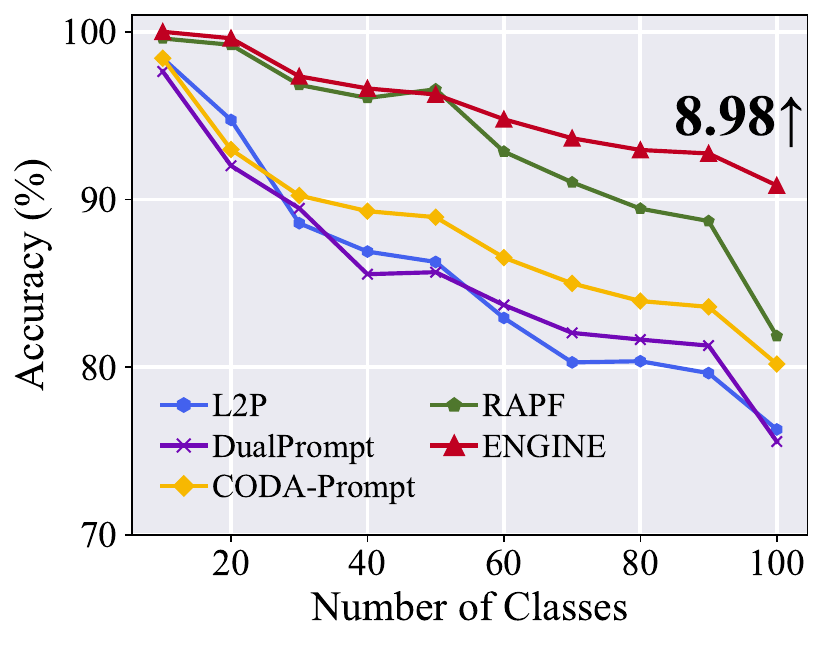}
	\caption{Experiments when using OpenAI weights on UCF B0 Inc10. {\bf \name consistently outperforms other methods with various backbone weights.}}
	\label{fig:supp-backbone}
\end{figure}

In the main paper, we mainly consider CLIP with ViT-B/16 under LAION400M pre-trained weight~\cite{ilharco_gabriel_2021_5143773}\footnote{https://github.com/mlfoundations/open\_clip} to conduct the experiments. In this section, we also provide the results with OpenAI pre-trained weight\footnote{https://github.com/openai/CLIP} on \ UCF B0 Inc10 in Figure~\ref{fig:supp-backbone}. As we can infer from the figure, \name consistently outperforms other methods with various backbone weights.

\subsection{Different LLMs}
\begin{figure}[h]
	\centering
	\includegraphics[width=0.8\columnwidth]{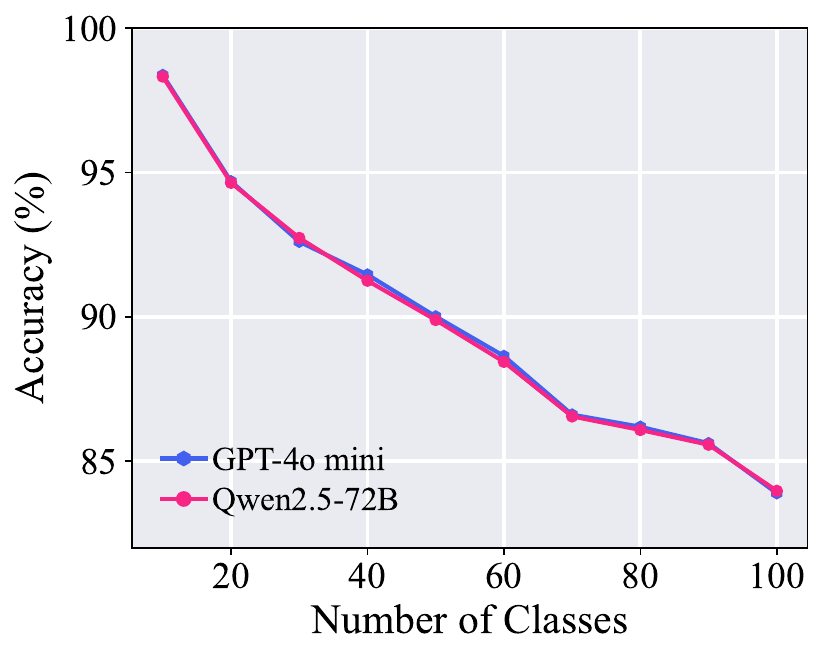}
	\caption{Results on Food B0 Inc10 with different LLMs for text description.  {\bf \name is robust and compatible with various LLMs. }}
	\label{fig:supp-llm}
\end{figure}

In the main paper, we mainly utilize GPT-4o mini to generate the class descriptions. Since \name is a general framework that is compatible with various LLMs, we consider Qwen2.5-72B~\cite{qwen} as the LLM to generate class descriptions, and conduct experiments on Food B0 Inc10. We report the comparison between GPT-4o mini and Qwen2.5-72B in Figure~\ref{fig:supp-llm}.

As shown in the figure, we find the performance results using different LLMs are quite similar, indicating the robustness of \name when using different LLMs.

\begin{figure*}
	\centering
	\begin{subfigure}{0.33\linewidth}
		\includegraphics[width=1\columnwidth]{pics/aircraft}
		\caption{Aircraft Base0 Inc10}
	\end{subfigure}
	\hfill
	\begin{subfigure}{0.33\linewidth}
		\includegraphics[width=1\linewidth]{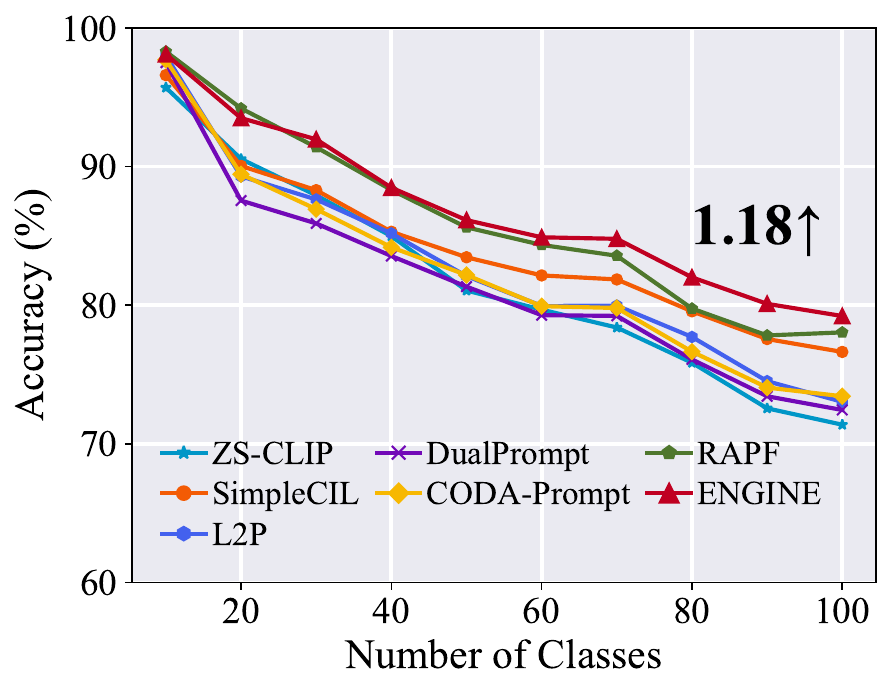}
		\caption{CIFAR100 Base0 Inc10}
	\end{subfigure}
	\hfill
	\begin{subfigure}{0.33\linewidth}
		\includegraphics[width=1\linewidth]{pics/cars}
		\caption{Cars Base0 Inc10}
	\end{subfigure}
	\\
	\begin{subfigure}{0.33\linewidth}
		\includegraphics[width=1\linewidth]{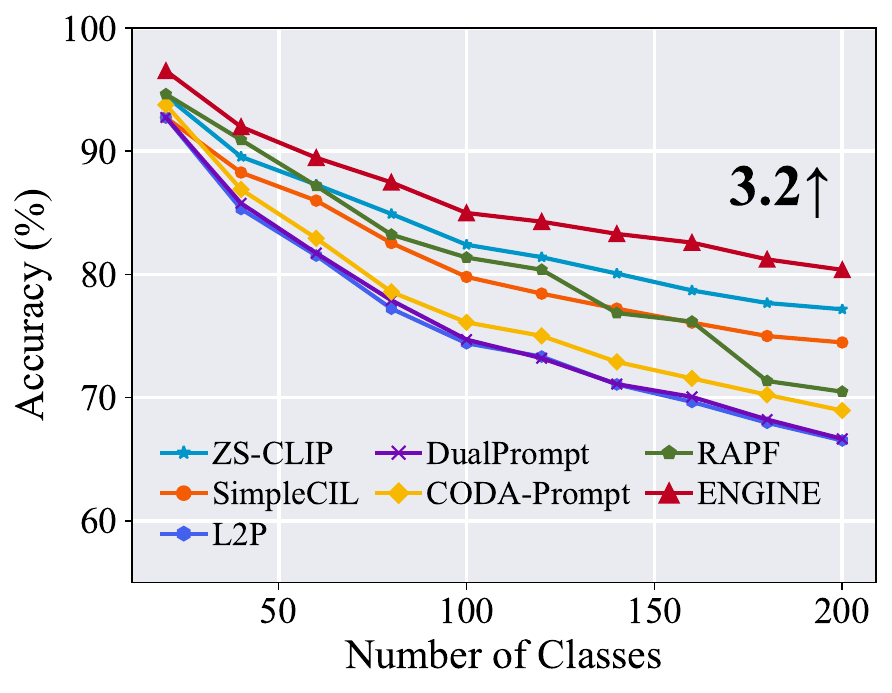}
		\caption{ImageNet-R Base0 Inc20}
	\end{subfigure}
	\hfill
	\begin{subfigure}{0.33\linewidth}
		\includegraphics[width=1\linewidth]{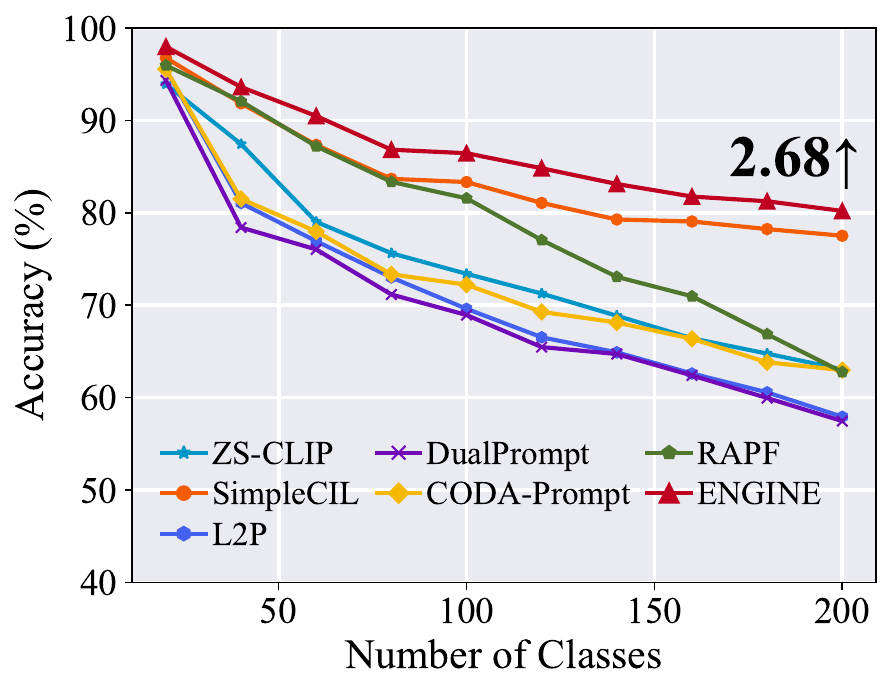}
		\caption{CUB Base0 Inc20}
	\end{subfigure}
	\hfill
	\begin{subfigure}{0.33\linewidth}
		\includegraphics[width=1\columnwidth]{pics/ucf}
		\caption{UCF Base0 Inc10}
	\end{subfigure}
	\\
	\begin{subfigure}{0.33\linewidth}
		\includegraphics[width=1\linewidth]{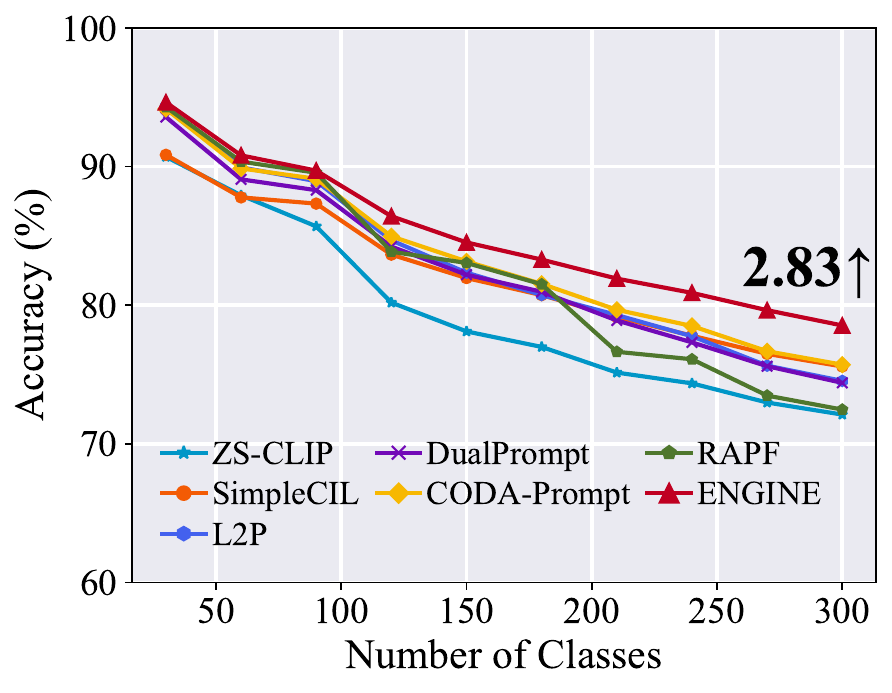}
		\caption{SUN Base0 Inc30}
	\end{subfigure}
	\hfill
	\begin{subfigure}{0.33\linewidth}
		\includegraphics[width=1\linewidth]{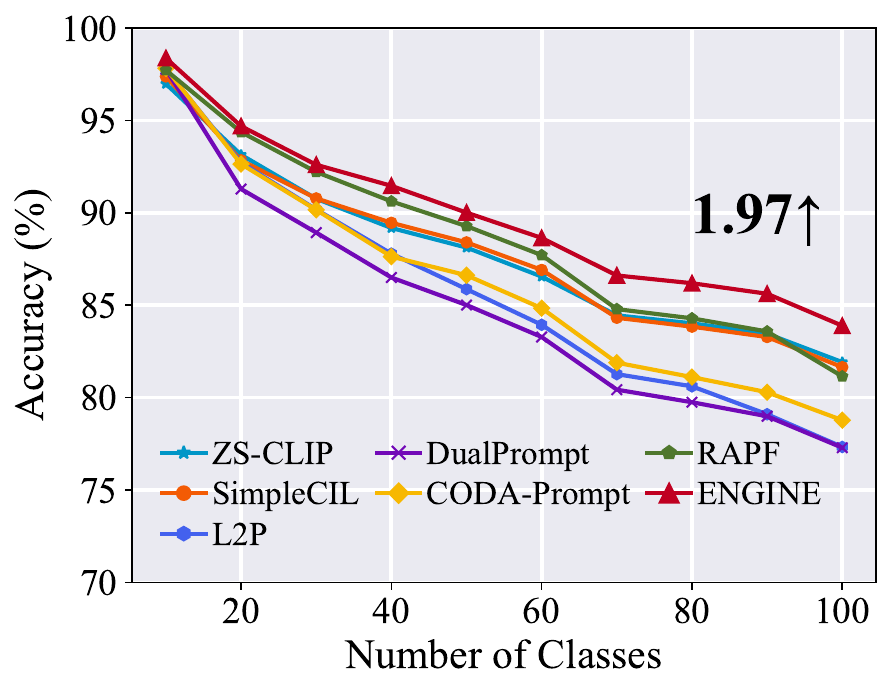}
		\caption{Food Base0 Inc10}
	\end{subfigure}
	\hfill
	\begin{subfigure}{0.33\linewidth}
		\includegraphics[width=1\columnwidth]{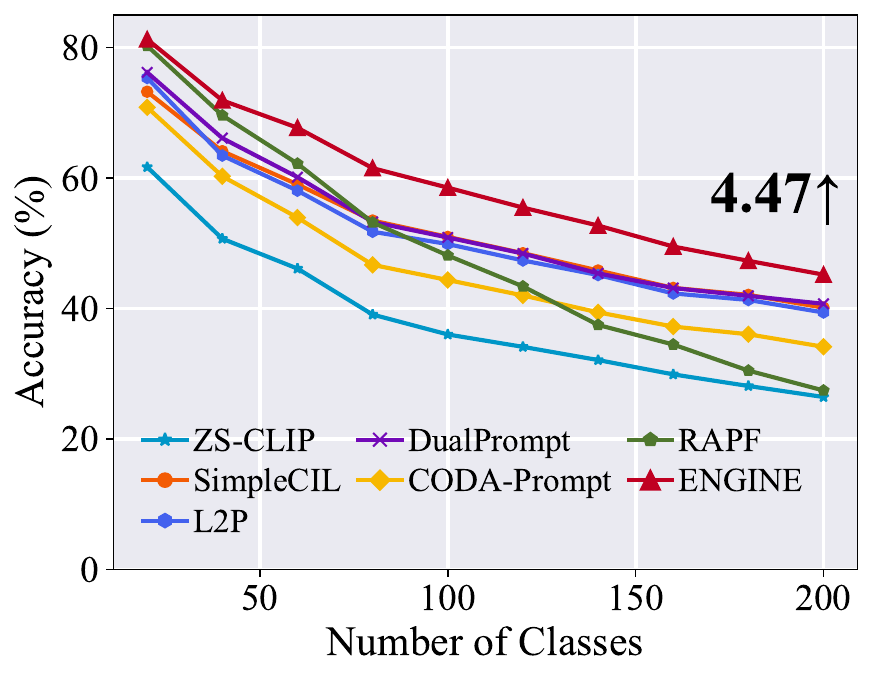}
		\caption{ObjectNet Base0 Inc20}
	\end{subfigure}
	\caption{ 	Incremental performance of different methods on B0 setting. We report the performance gap after the last incremental stage of \name and the runner-up method at the end of the line.    All methods utilize the same CLIP pre-trained weight. }
	\label{fig:supp-benchmark-b0}
\end{figure*}

\begin{figure*}
	\centering
	\begin{subfigure}{0.33\linewidth}
		\includegraphics[width=1\columnwidth]{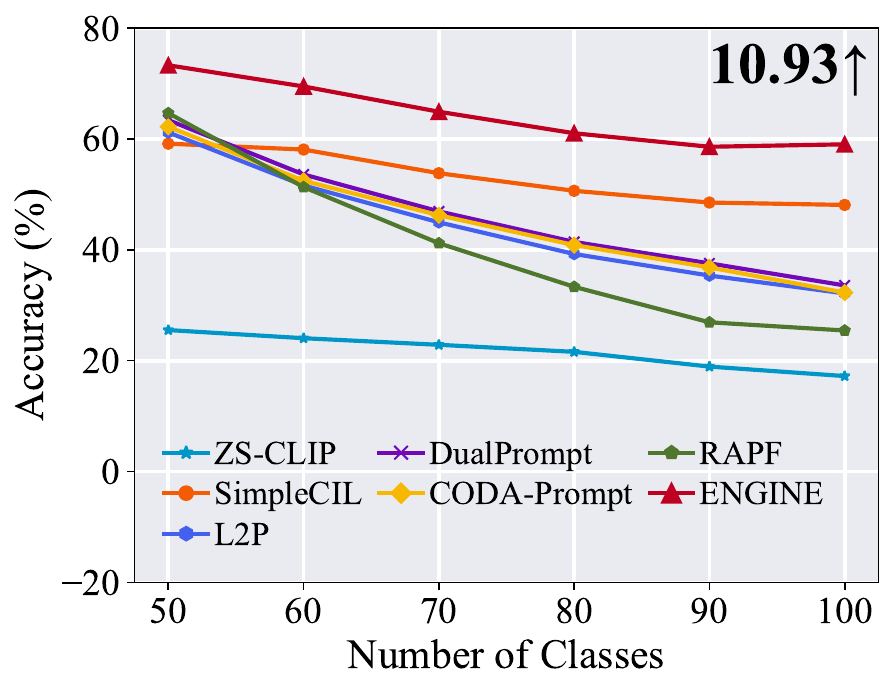}
		\caption{Aircraft Base50 Inc10}
		\label{fig:benchmark-aircraft50}
	\end{subfigure}
	\hfill
	\begin{subfigure}{0.33\linewidth}
		\includegraphics[width=1\linewidth]{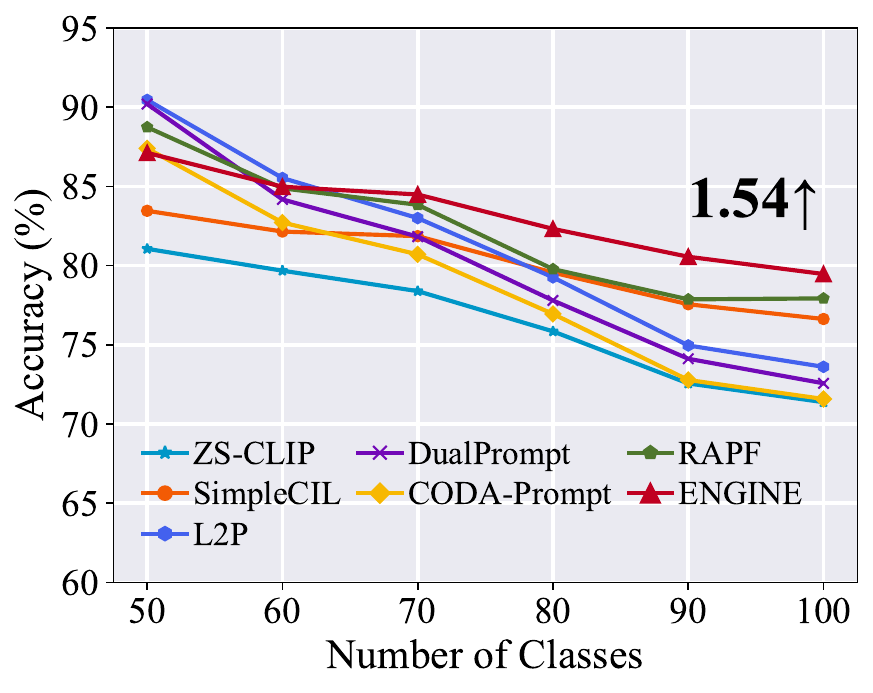}
		\caption{CIFAR100 Base50 Inc10}
		\label{fig:benchmark-cifar50}
	\end{subfigure}
	\hfill
	\begin{subfigure}{0.33\linewidth}
		\includegraphics[width=1\linewidth]{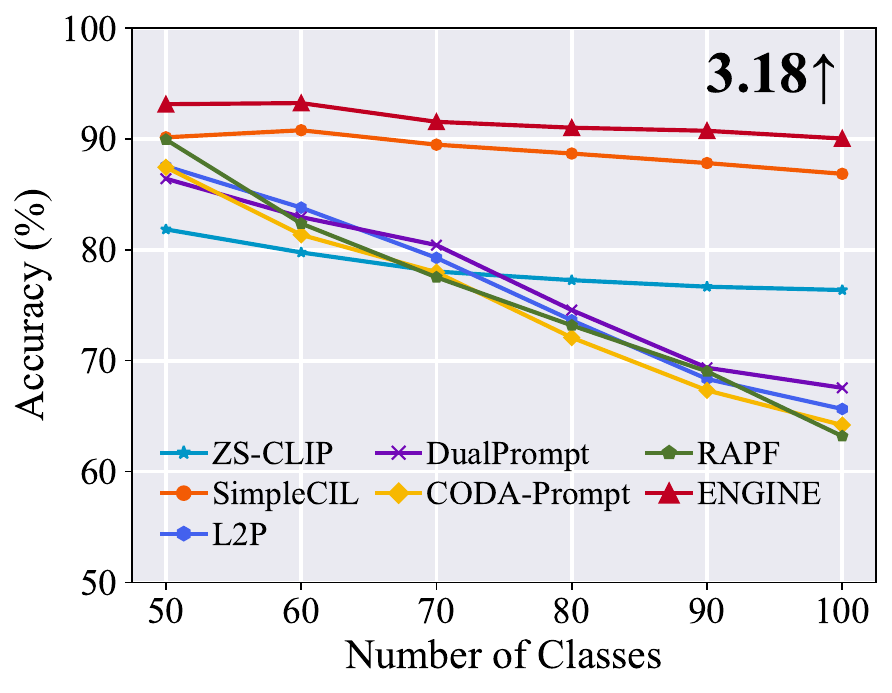}
		\caption{Cars Base50 Inc10}
		\label{fig:benchmark-cars50}
	\end{subfigure}
	\\
	\begin{subfigure}{0.33\linewidth}
		\includegraphics[width=1\linewidth]{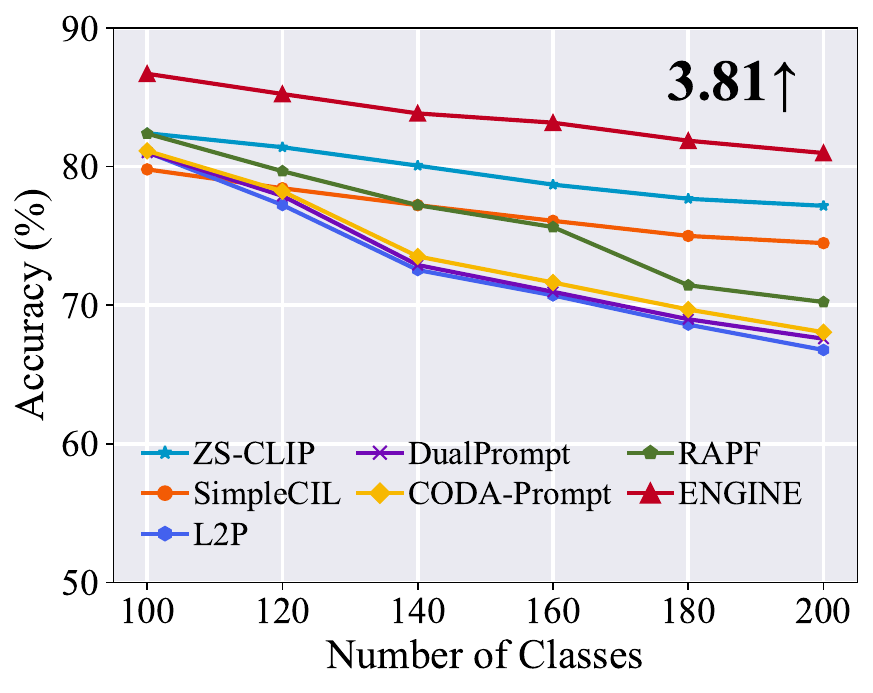}
		\caption{ImageNet-R Base100 Inc20}
		\label{fig:benchmark-imagenetr100}
	\end{subfigure}
	\hfill
	\begin{subfigure}{0.33\linewidth}
		\includegraphics[width=1\linewidth]{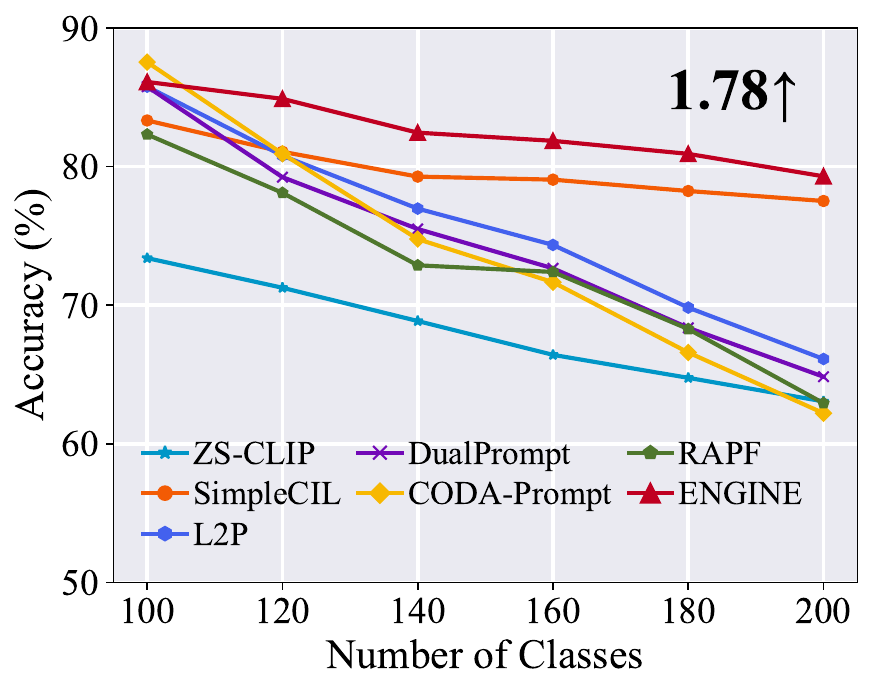}
		\caption{CUB Base1000 Inc20}
		\label{fig:benchmark-cub100}
	\end{subfigure}
	\hfill
	\begin{subfigure}{0.33\linewidth}
		\includegraphics[width=1\columnwidth]{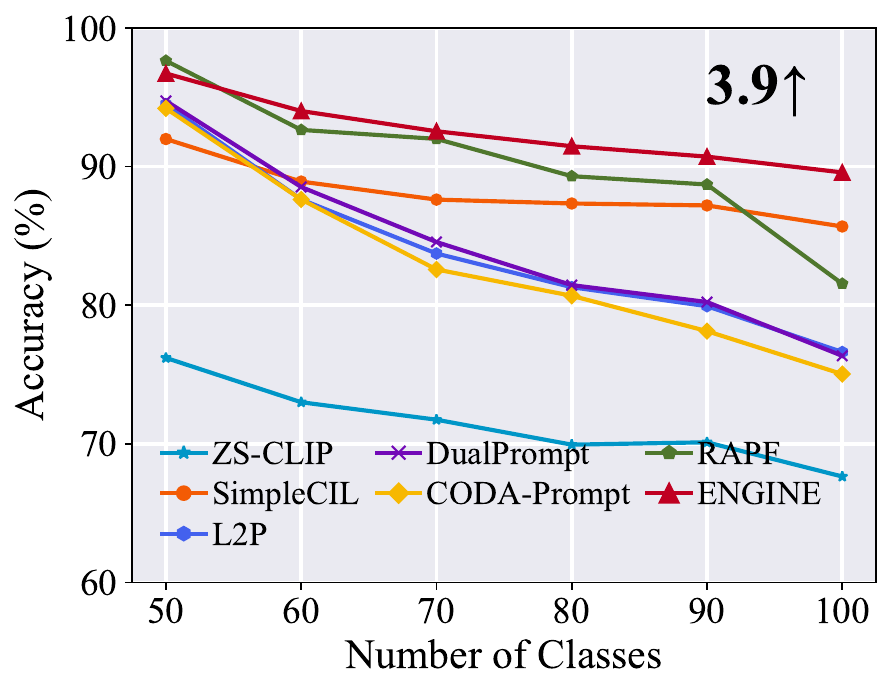}
		\caption{UCF Base50 Inc10}
		\label{fig:benchmark-ucf50}
	\end{subfigure}
	\\
	\begin{subfigure}{0.33\linewidth}
		\includegraphics[width=1\linewidth]{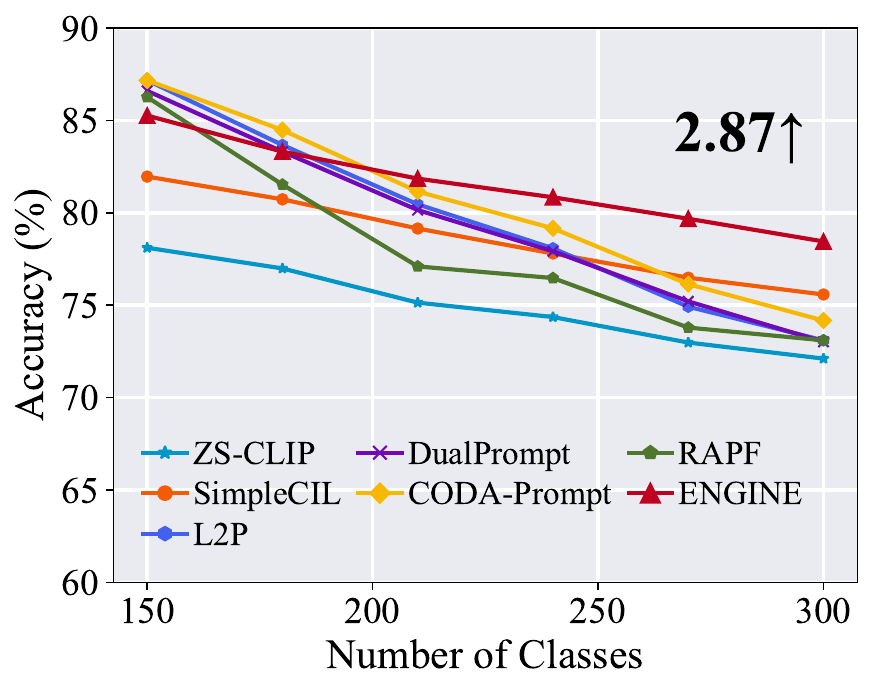}
		\caption{SUN Base150 Inc30}
		\label{fig:benchmark-sun150}
	\end{subfigure}
	\hfill
	\begin{subfigure}{0.33\linewidth}
		\includegraphics[width=1\linewidth]{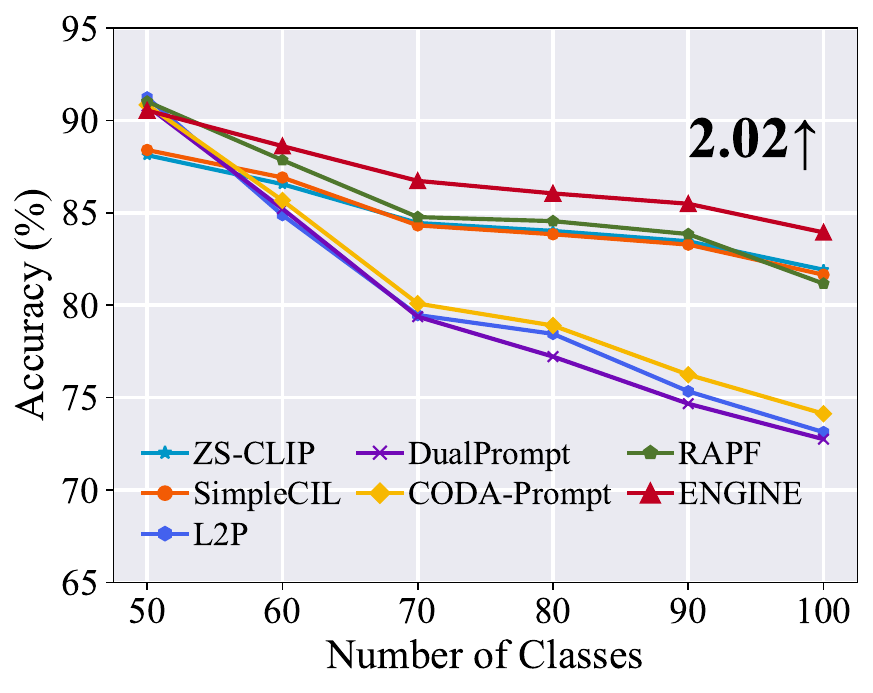}
		\caption{Food Base50 Inc10}
		\label{fig:benchmark-food50}
	\end{subfigure}
	\hfill
	\begin{subfigure}{0.33\linewidth}
		\includegraphics[width=1\columnwidth]{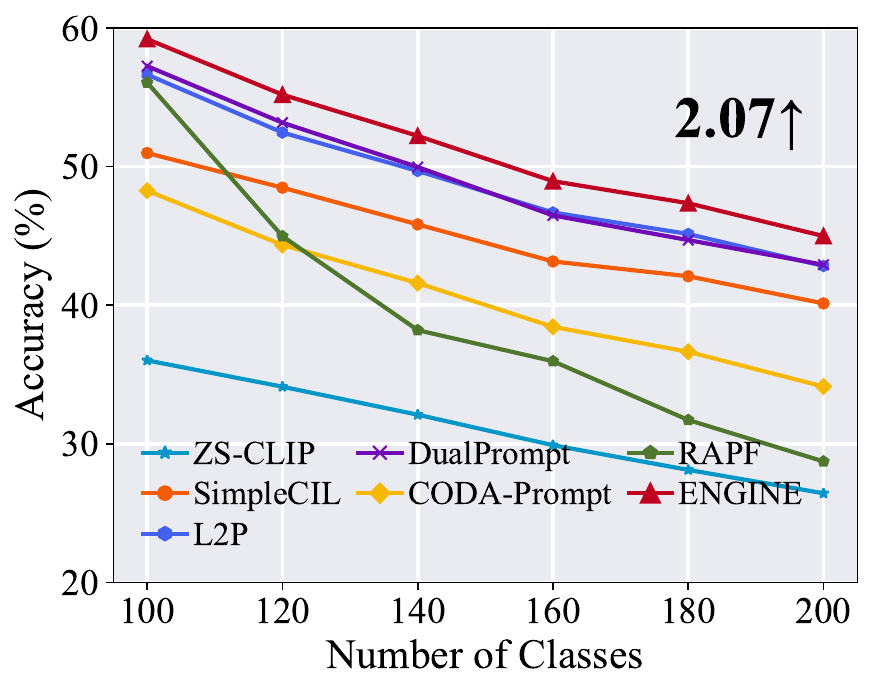}
		\caption{ObjectNet Base100 Inc20}
		\label{fig:benchmark-objectnet100}
	\end{subfigure}
	\caption{ 	Incremental performance of different methods on half-base setting. We report the performance gap after the last incremental stage of \name and the runner-up method at the end of the line.    All methods utilize the same CLIP pre-trained weight. }
	\label{fig:supp-benchmark-b50}
\end{figure*}

\begin{figure*}[!h]
	\centering
	{\includegraphics[width=.99\columnwidth]{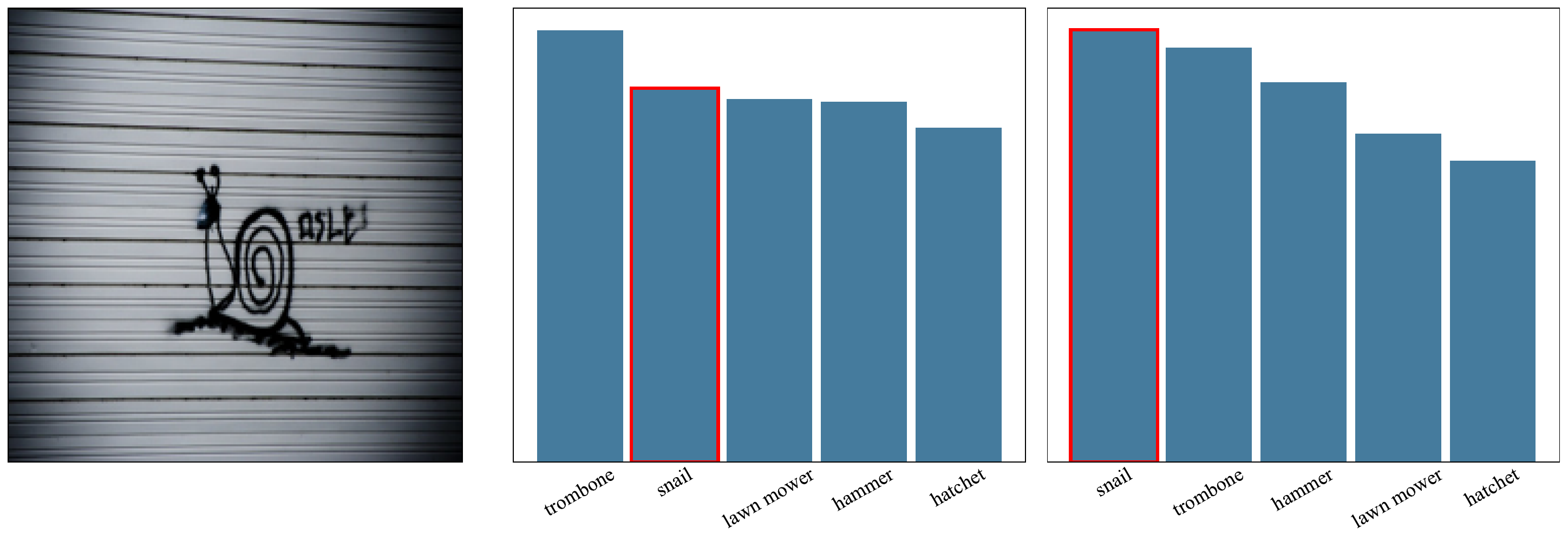}}
	{\includegraphics[width=.99\columnwidth]{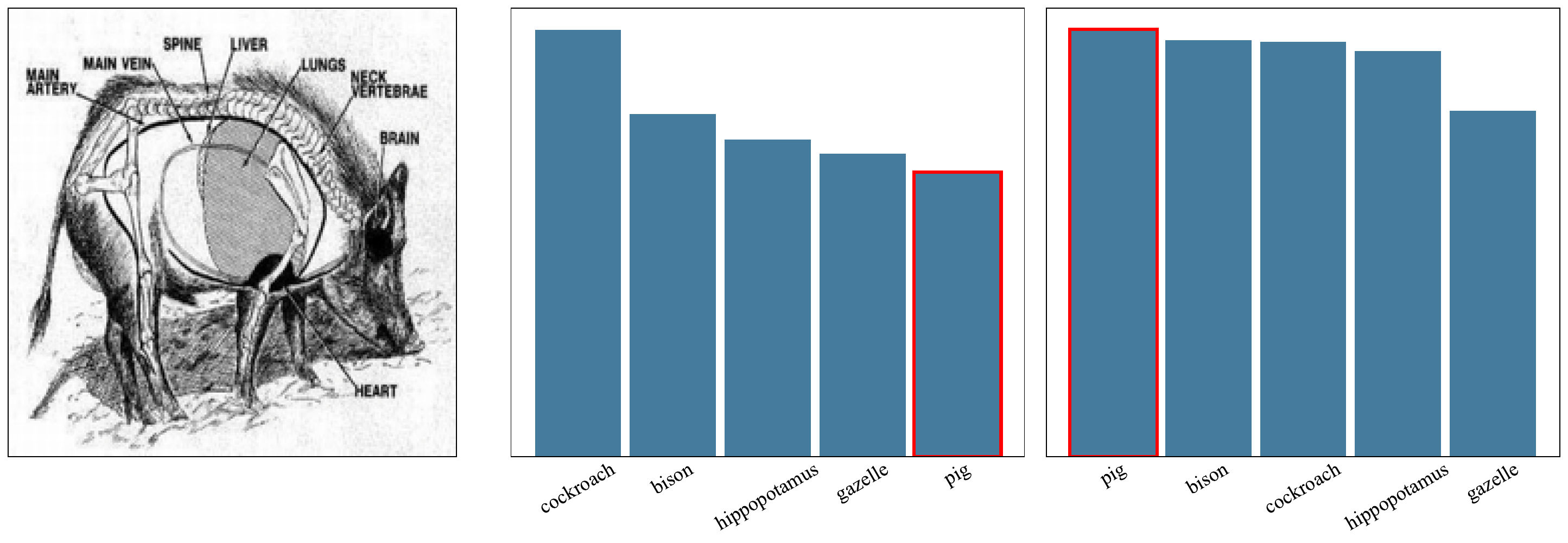}}
	{\includegraphics[width=.99\columnwidth]{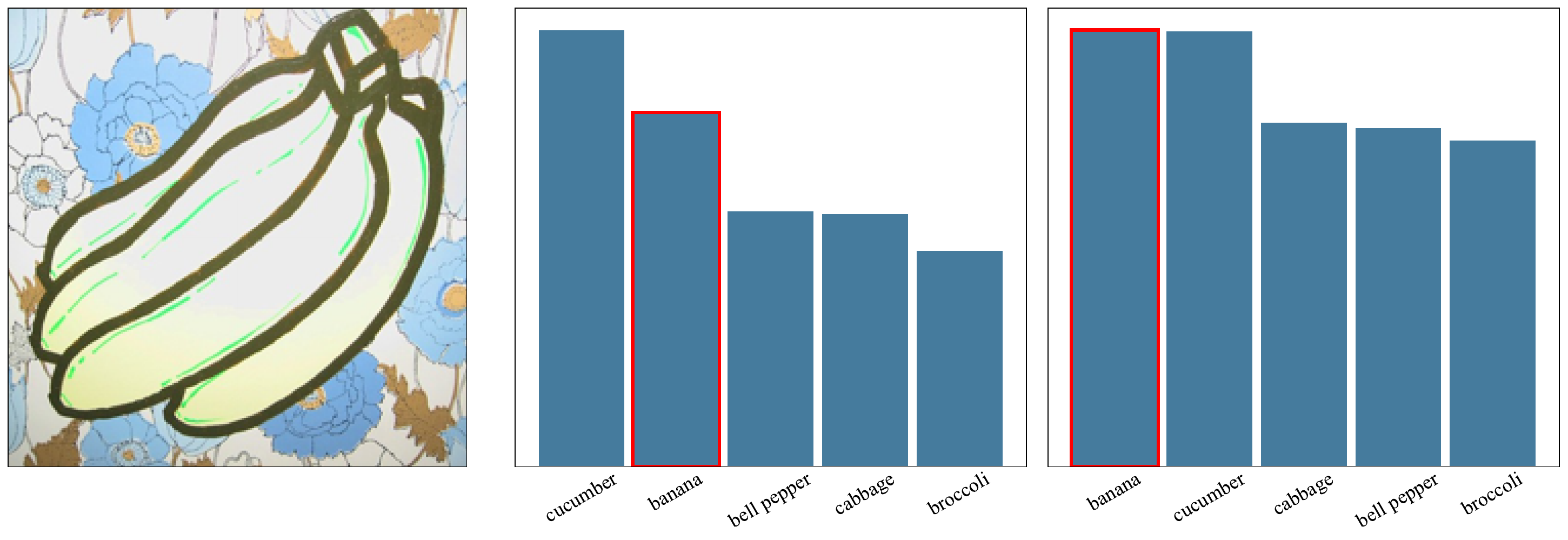}}
	{\includegraphics[width=.99\columnwidth]{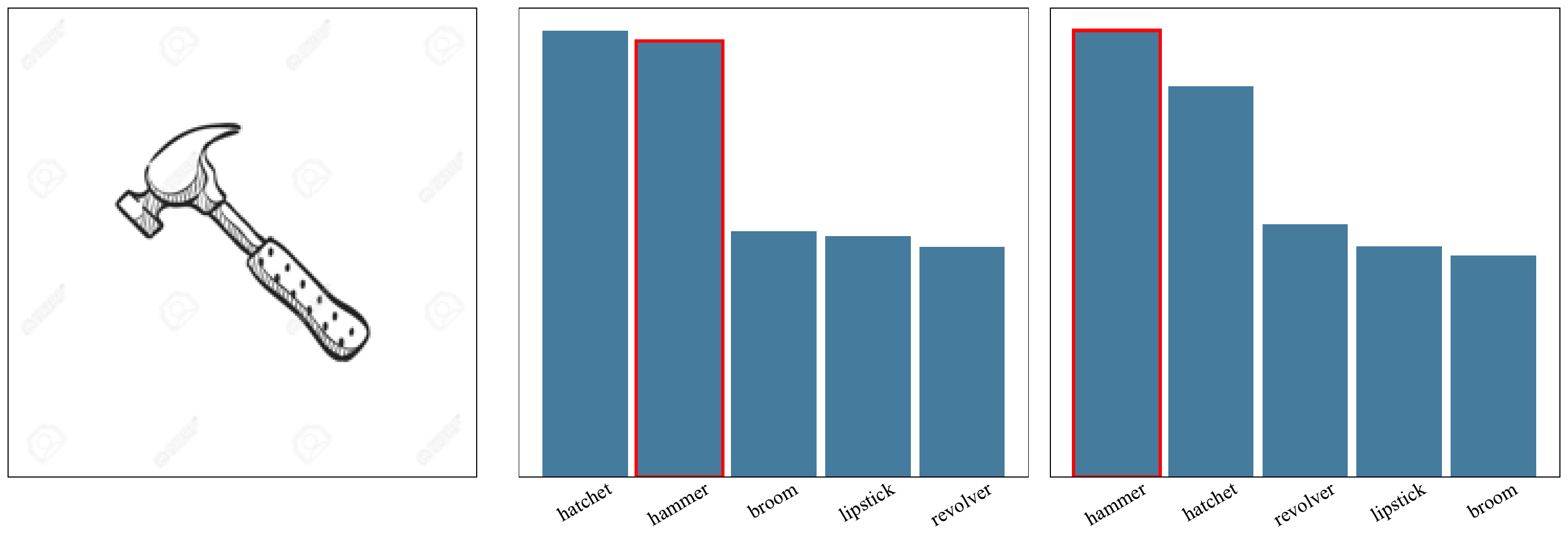}}
	{\includegraphics[width=.99\columnwidth]{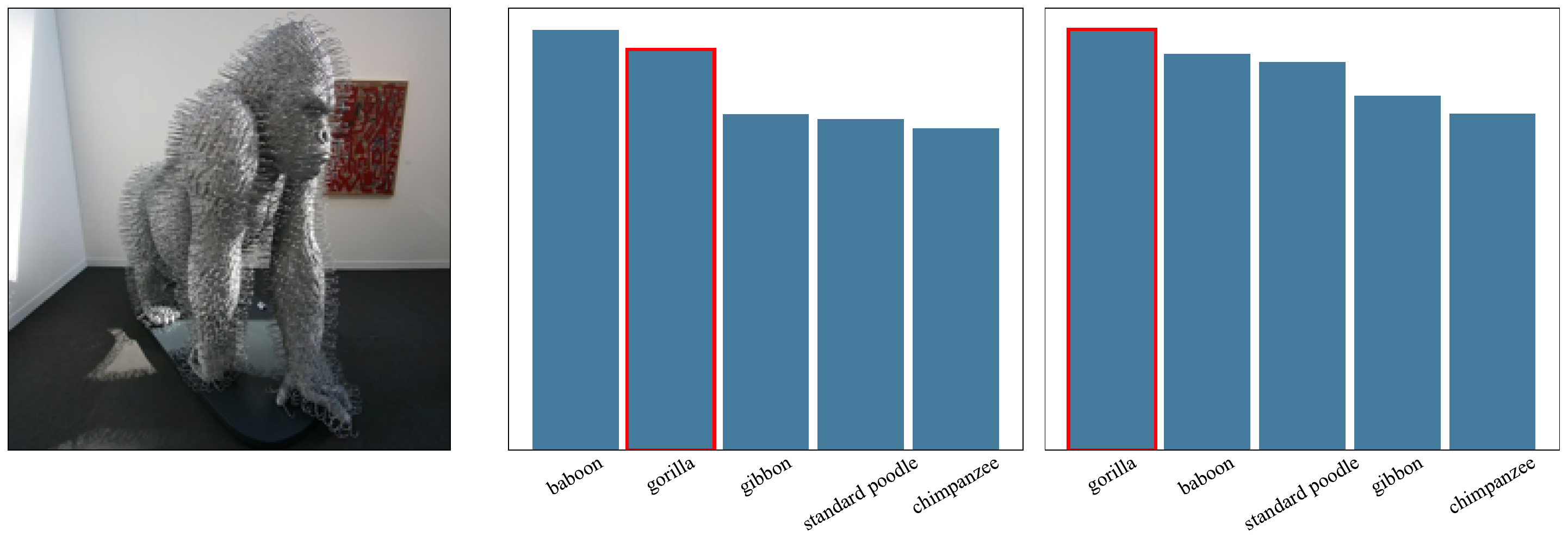}}
	{\includegraphics[width=.99\columnwidth]{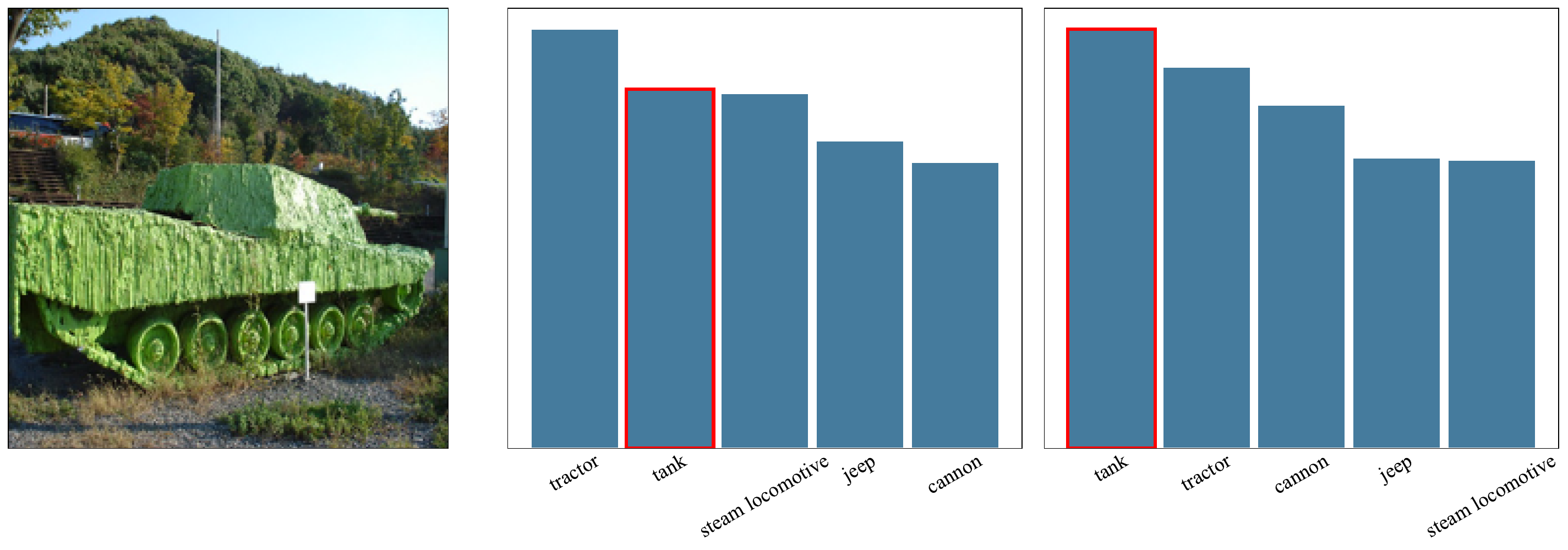}}
	{\includegraphics[width=.99\columnwidth]{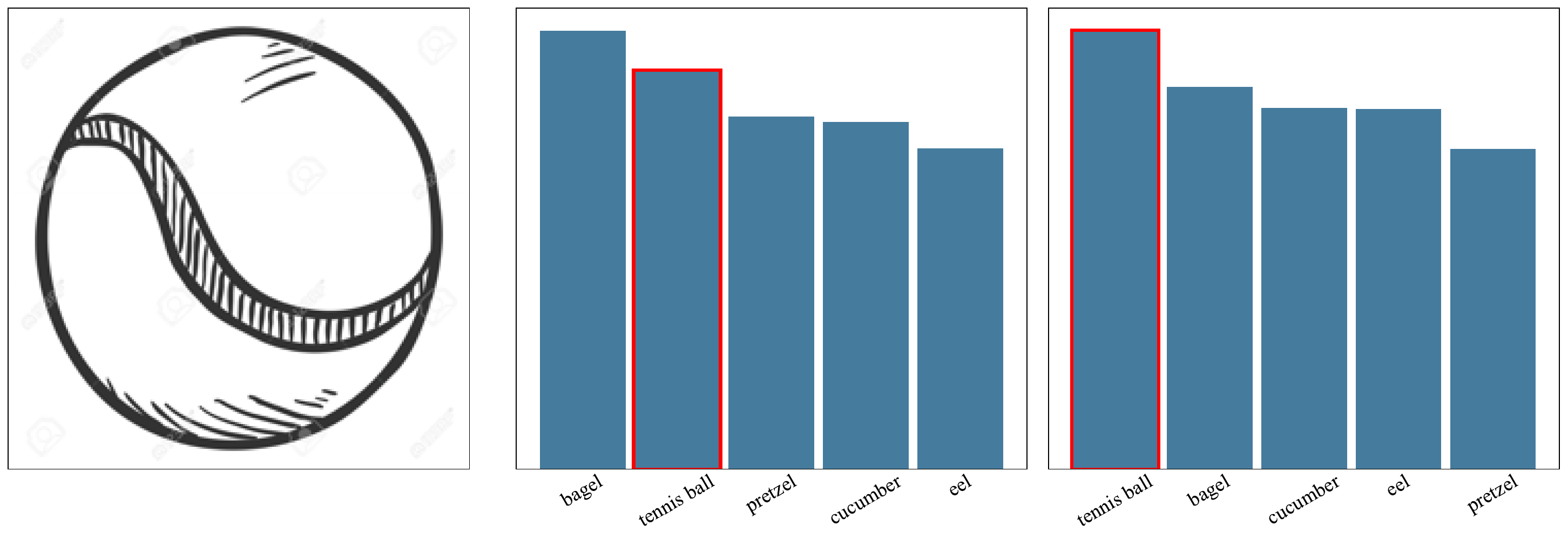}}
	{\includegraphics[width=.99\columnwidth]{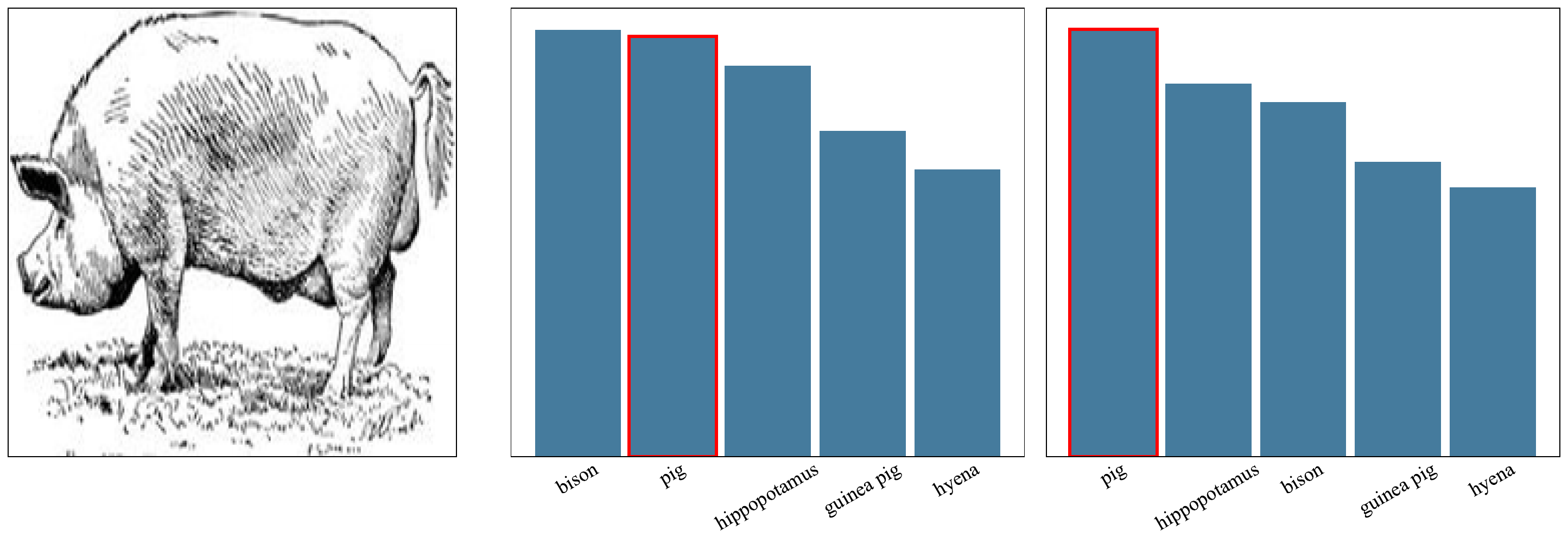}}
	{\includegraphics[width=.99\columnwidth]{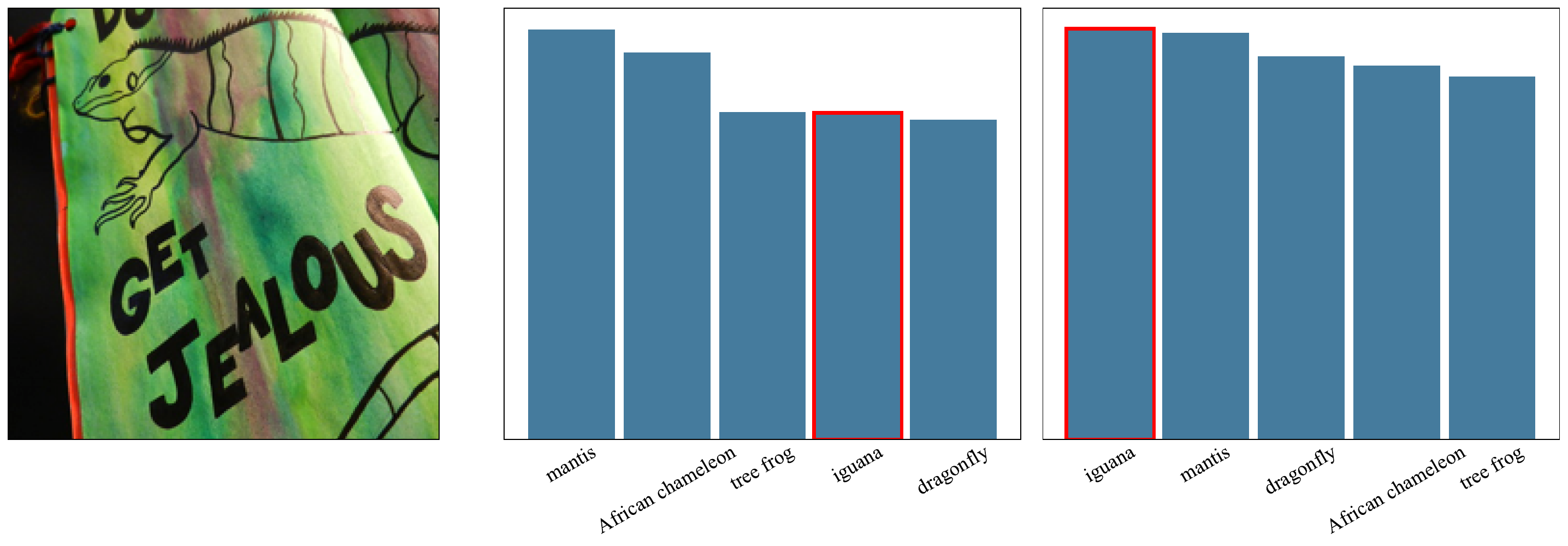}}
	{\includegraphics[width=.99\columnwidth]{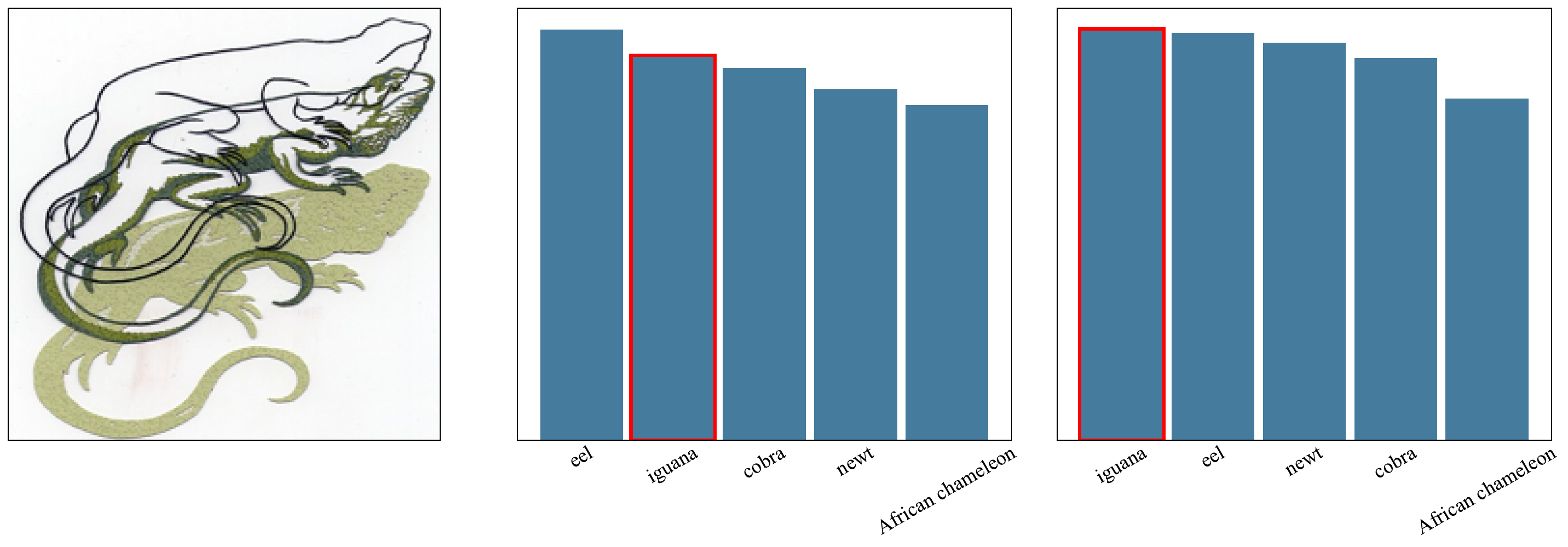}}
	{\includegraphics[width=.99\columnwidth]{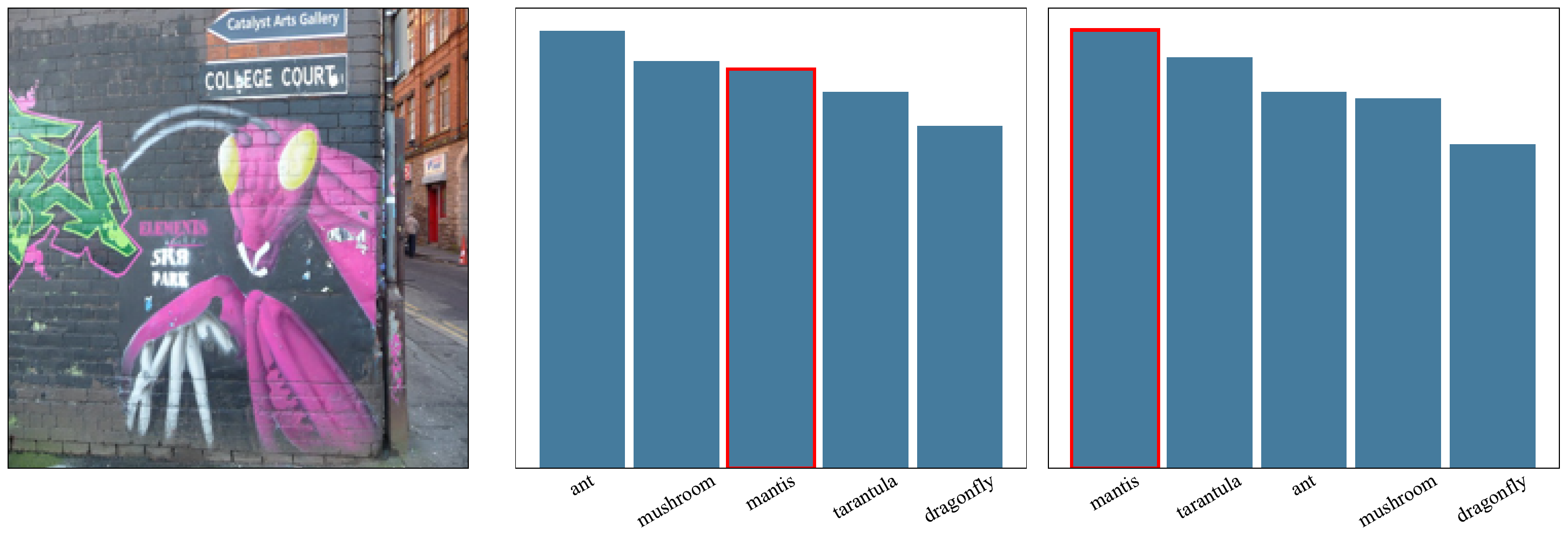}}
	{\includegraphics[width=.99\columnwidth]{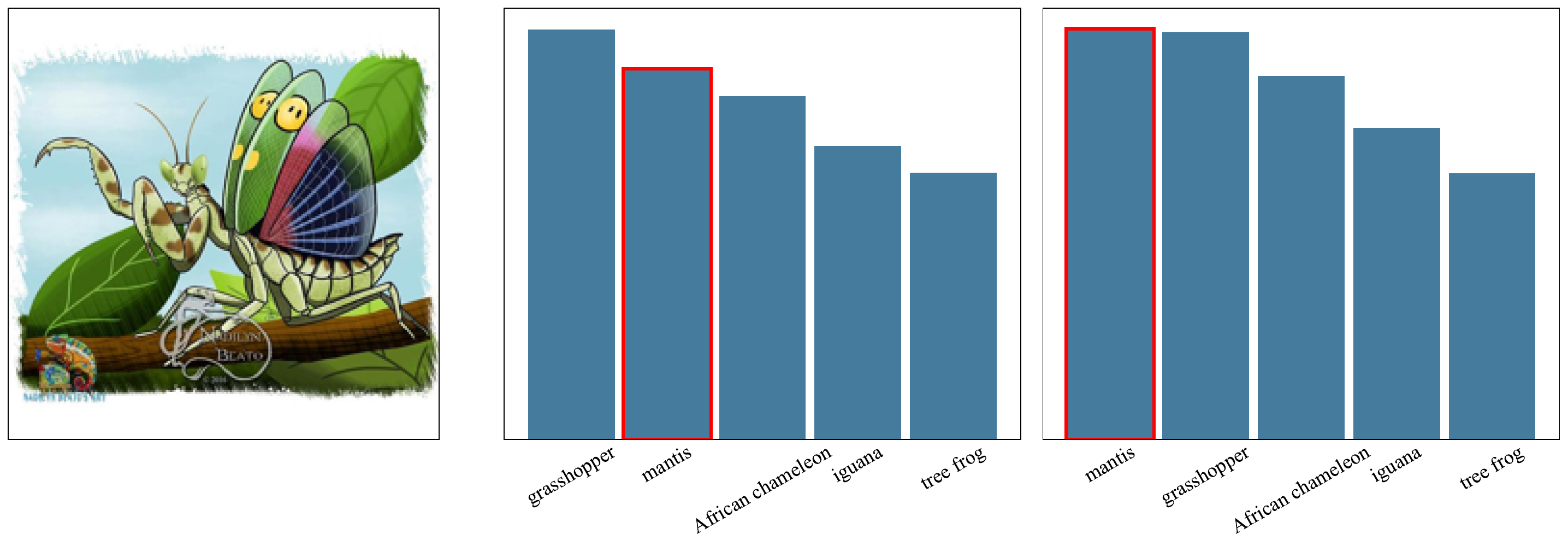}}
	{\includegraphics[width=.99\columnwidth]{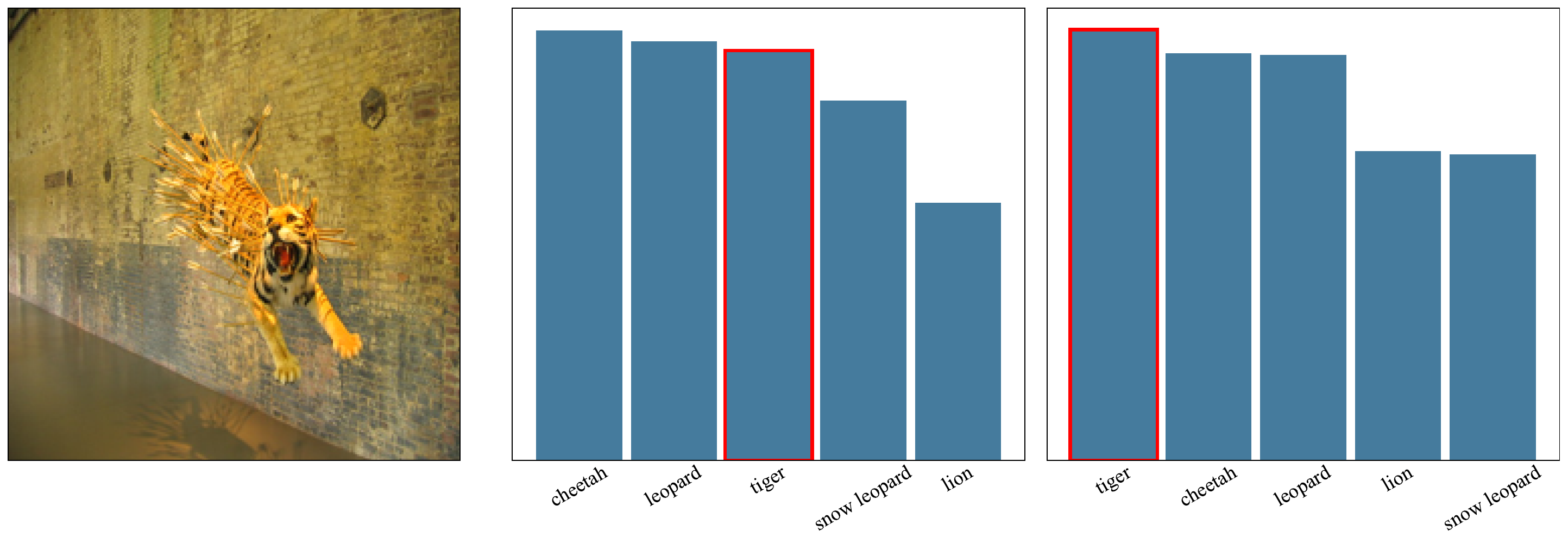}}
	{\includegraphics[width=.99\columnwidth]{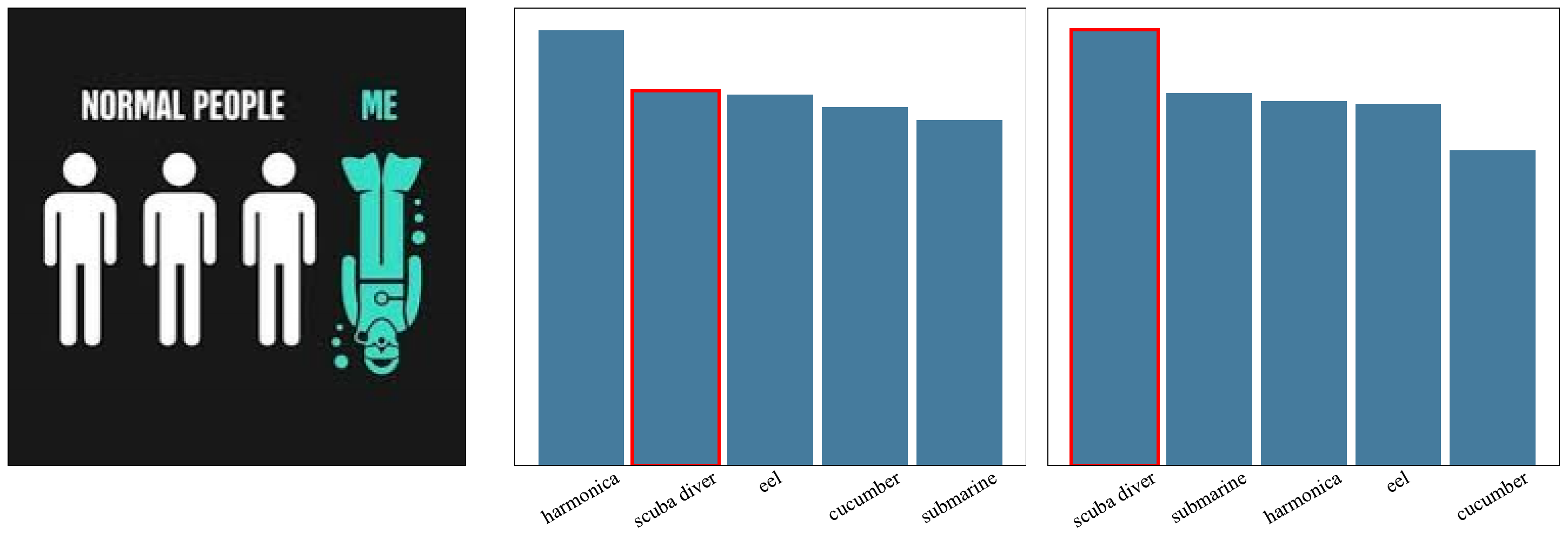}}
	\caption{ {\bf Left:} Input. {\bf Middle:} Top-5 predictions before post-tuning knowledge injection.  {\bf Right:} Top-5 predictions after post-tuning. More cases are shown in the supplementary.}
	\label{fig:supp-further-analysis}
\end{figure*}

\section{Introduction About Compared Methods}\label{sec:supp_intro}

In this section, we introduce the details of the compared methods adopted in the main paper. For a fair comparison, all methods are based on the same pre-trained model. The details of the compared methods in Table 1 are listed as:

\begin{itemize}
	\item {\bf Finetune}: with a pre-trained CLIP as initialization, it finetunes CLIP for every new task. Hence, it suffers severe catastrophic forgetting on former tasks.
	\item \textbf{CoOp~\cite{zhou2022learning}:} This approach freezes both the image encoder and text encoder of the pre-trained CLIP. It optimizes a learnable prompt tensor $\mathbf{t}$ using contrastive loss.
	\item \textbf{SimpleCIL~\cite{zhou2023revisiting}:} This method relies on the pre-trained image encoder and does not involve the text encoder. Hence, in the pre-trained CLIP, we drop the text branch and only use the visual branch for evaluation. The frozen image encoder extracts class centers (prototypes) for each new class, and a cosine classifier is utilized for classification. Since the model is not updated via backpropagation, it showcases the generalizability of the pre-trained vision encoder on downstream tasks.		\item \textbf{ZS-CLIP~\cite{radford2021learning}:} This baseline freezes the pre-trained CLIP and predicts the logits of each incoming class using cosine similarity. It serves as a reference for the performance of pre-trained CLIP on downstream tasks.
	\item {\bf L2P~\cite{wang2022learning}}: This method only involves the visual branch of CLIP. During model updating, it freezes the pre-trained weights and utilizes visual prompt tuning~\cite{jia2022visual} to trace the new task's features. It builds instance-specific prompts with a prompt pool, which is constructed via key-value mapping.  
	\item  {\bf DualPrompt~\cite{wang2022dualprompt}}: is an extension of L2P, which extends the prompt into two types, \ie, general and expert prompts. The other details are kept the same with L2P, \ie, using the prompt pool to build instance-specific prompts. This method only involves the visual branch of CLIP.
	\item {\bf CODA-Prompt~\cite{smith2023coda}}: noticing the drawback of instance-specific prompt selection, it aims to eliminate the prompt selection process by prompt reweighting. The prompt selection process is replaced with an attention-based prompt recombination. This method only involves the visual branch of CLIP.
	\item {\bf RAPF~\cite{huang2024class}}: aims to enhance the continual learning ability of CLIP. It combines the hard class separation loss and decomposed parameter fusion to encode new knowledge into the CLIP model.
	
\end{itemize}

The above methods are exemplar-free, which do not require using exemplars. We also compare some exemplar-based methods in Table 2 of the main paper as follows:

\begin{itemize}
	
	\item {\bf iCaRL~\cite{rebuffi2017icarl}}: This method only involves the visual branch of CLIP. It utilizes knowledge distillation and exemplar replay to recover former knowledge. It also utilizes the nearest center mean classifier for final classification.
	\item {\bf MEMO~\cite{zhou2022model}}: This method only involves the visual branch of CLIP.  It decouples the network structure into specialized (deep) and generalized (shallow) layers and extends specialized layers based on the shared generalized layers. Hence, the memory cost for network expansion decreases from a whole backbone to generalized blocks. In the implementation, we follow~\cite{zhou2022model} to decouple the vision transformer at the last transformer block.
	\item {\bf PROOF~\cite{zhou2023learning}}: aims to enhance CLIP's continual learning ability by learning expandable projection layers and cross-modal fusion module. The prototypes of historical visual and textual features are passed through the cross-modal fusion for further matching.
	
\end{itemize}

In the experiments, we reimplement the above methods based on their source code and PyCIL\footnote{https://github.com/G-U-N/PyCIL}.

\section{Full Results} \label{sec:supp_full_results}

In this section, we show more experimental results of different methods. 
In the main paper, we only report three typical learning trends among compared methods. In this section, we report the full results corresponding to Table 1 of the main paper. Specifically, we report the incremental performance of different methods with 0 base classes in Figure~\ref{fig:supp-benchmark-b0} and half base classes in Figure~\ref{fig:supp-benchmark-b50}. 
As shown in these results, \name consistently outperforms other methods on different datasets and different data splits by a substantial margin.

\section{More Visualizations} \label{sec:supp_vis}
\subsection{Adjusted predictions}
In this section, we provide more visualizations of the effect of post-tuning knowledge injection in Figure~\ref{fig:supp-further-analysis}. As shown in the figure,  post-tuning efficiently adjusts the predictions of the model to highlight the ground truth class.

\subsection{Generated descriptions}

In this section, we give examples of the generated descriptions of class-wise discriminative features. We give the examples on CIFAR100:

\blue{Camel versus cattle}:
\begin{displayquote}
	{\bf Q}: What are unique visual features of camel compared to cattle in a photo? Focus on their key visual differences. \\
	{\bf A}: Camel: {\bf 1.} Tall, humped back which distinguishes it from cattle. {\bf 2.} Long legs and a proportionally longer neck, giving a different silhouette. {\bf 3.} Typically covered in coarse fur, often light brown or beige.  {\bf 4.} Feet have padding for desert terrains, contrasting with cattle's hooves. {\bf 5.}Can often be seen with a distinctive gait, differing from the more sturdy stance of cattle.  \\ Cattle: {\bf 1.} Large, bulky bodies with short legs. {\bf 2.} Varying coat colors from black to brown to white. {\bf 3.} Prominent horns on some breeds, depending on gender.  {\bf 4.} Large, expressive eyes with long eyelashes.  {\bf 5.} Commonly seen grazing in fields or pastures.
\end{displayquote}

\blue{Shark versus ray}:
\begin{displayquote}
	{\bf Q}: What are unique visual features of shark compared to ray in a photo? Focus on their key visual differences. \\
	{\bf A}: Shark: {\bf 1.}Streamlined body with a pointed snout and sharp teeth.  {\bf 2.} A dorsal fin prominently displayed on the top.{\bf 3.} Pelvic fins and pectoral fins that assist in maneuvering. {\bf 4.}Typically gray or blueish skin, often with a lighter belly. {\bf 5.} Can be shown swimming in a sleek, powerful posture. \\ Ray: {\bf 1.} Flattened body shape with a broad, disc-like structure.  {\bf 2.} Long, pointed snout extending from the disc. {\bf 3.} Wing-like pectoral fins that extend outwards. {\bf 4.} Short tail that may have a spine or barb.  {\bf 5.} Coloration often varying from gray to brown.
\end{displayquote}

\blue{Bed versus couch}:
\begin{displayquote}
	{\bf Q}: What are unique visual features of bed compared to couch in a photo? Focus on their key visual differences. \\
	{\bf A}: Bed: {\bf 1.}Typically rectangular shape, varying in size.  {\bf 2.} Soft, cushioned surface with a mattress.{\bf 3.} Often has a headboard or footboard. {\bf 4.}May include bedding such as blankets and pillows. {\bf 5.} Primarily found indoors in a home setting. \\ Couch: {\bf 1.} Large, upholstered furniture piece with a soft seating area.  {\bf 2.} Typically consists of cushions that vary in firmness. {\bf 3.} Often rectangular or L-shaped in design. {\bf 4.} Can have armrests and back support, with decorative fabric.  {\bf 5.} Used in living rooms or lounges, contrasting a flat surface.
\end{displayquote}

We also give the examples on ImageNet-R:

\blue{Harp versus electric guitar}:

\begin{displayquote}
	{\bf Q}: What are unique visual features of harp compared to electric guitar in a photo? Focus on their key visual differences. \\
	{\bf A}:Harp: {\bf 1.}Large, triangular frame with a curved top and vertical strings.  {\bf 2.} Tall and elegant, often made of wood.{\bf 3.} Strings are arranged in a straight line from top to bottom, creating tension. {\bf 4.}Pedals at the base for changing pitch on certain types. {\bf 5.} Generally features ornate carvings and a polished finish. \\ Electric guitar: {\bf 1.} Solid body with a distinctive curvy or pointed shape. {\bf 2.} Strings visible on top along a long neck, attached to a circular body. {\bf 3.}Often has pickups, knobs, and frets, creating a complex visual profile. {\bf 4.} Typically made of wood with a shiny finish and various colors. {\bf 5.} Centered around sound production through electrical means unlike the strings of a harp that are different in construction.
\end{displayquote}

\blue{Pizza versus burrito}:
\begin{displayquote}
	{\bf Q}: What are unique visual features of pizza compared to burrito in a photo? Focus on their key visual differences. \\
	{\bf A}: Pizza: {\bf 1.}Flat, round shape with a raised crust along the perimeter.  {\bf 2.} Visible layers of toppings, sauce, and cheese creating a colorful appearance.{\bf 3.} Often has a glossy sheen from melted cheese, unlike the matte finish of a burrito. {\bf 4.}Slices that can show a variety of ingredients and color contrasts. {\bf 5.} No wrapping it is served open-faced, showcasing toppings, unlike a burrito. \\ Burrito: {\bf 1.} Cylindrical shape wrapped in a soft, flat tortilla. {\bf 2.} Often showcases a colorful mixture of fillings (rice, beans, meat). {\bf 3.}Typically closed at both ends, contrasting with pizza open slices. {\bf 4.} Appears larger and thicker than a typical slice of pizza.  {\bf 5.} Topping textures hidden within, rather than spread across the surface.
\end{displayquote}

\end{document}